\documentclass[a4paper,onecolumn,accepted=2024-05-28]{quantumarticle}
\pdfoutput=1

\usepackage[numbers,sort&compress]{natbib}
\usepackage{graphicx}
\usepackage{amsmath}
\usepackage{epsfig}
\usepackage{amssymb}
\usepackage[utf8]{inputenc}
\usepackage[english]{babel}
\usepackage[T1]{fontenc}

\usepackage{xcolor}
  \definecolor{codegreen}{rgb}{0,0.6,0}
  \definecolor{codegray}{rgb}{0.5,0.5,0.5}
  \definecolor{codepurple}{rgb}{0.58,0,0.82}
  \definecolor{backcolour}{rgb}{0.95,0.95,0.92}
  \definecolor{purple}{RGB}{128,0,128}
\usepackage[colorlinks=true,citecolor=black,linkcolor=black,urlcolor=purple]{hyperref}
\usepackage{cleveref}

\usepackage{listings}
  \lstdefinestyle{mystyle}{
    language=Python,
    backgroundcolor=\color{backcolour},   
    commentstyle=\color{codegreen},
    keywordstyle=\color{magenta},
    numberstyle=\tiny\color{codegray},
    stringstyle=\color{codepurple},
    basicstyle=\ttfamily\footnotesize,
    breakatwhitespace=false,
    breaklines=true,
    captionpos=b,
    keepspaces=true,
    numbersep=5pt,
    showspaces=false,
    showstringspaces=false,
    showtabs=false,
    tabsize=2,
    escapechar=\%
}
\usepackage{tikzit}

\tikzstyle{leaf Node}=[fill={rgb,255: red,171; green,196; blue,255}, draw=black, shape=circle, tikzit category=Node, minimum width=0.6cm, minimum height=0.6cm]
\tikzstyle{data Node}=[fill={rgb,255: red,255; green,143; blue,171}, draw=black, shape=circle, tikzit category=Node, minimum width=0.6cm, minimum height=0.6cm]
\tikzstyle{virtual Node}=[fill=white, draw=black, shape=circle, tikzit category=Node, dashed, minimum width=0.6cm, minimum height=0.6cm]
\tikzstyle{resultant Node}=[fill={rgb,255: red,132; green,220; blue,198}, draw=black, shape=circle, tikzit category=Node, minimum width=0.6cm, minimum height=0.6cm]
\tikzstyle{leaf ParamNode}=[fill={rgb,255: red,171; green,196; blue,255}, draw=black, shape=rectangle, rounded corners=1pt, tikzit category=ParamNode, minimum width=0.6cm, minimum height=0.6cm]
\tikzstyle{virtual ParamNode}=[fill=white, draw=black, shape=rectangle, tikzit category=ParamNode, rounded corners=1pt, dashed, minimum width=0.6cm, minimum height=0.6cm]

\tikzstyle{Edge}=[-]
\tikzstyle{dashed_edge}=[-, dashed]

\newcommand{\tk}{TensorKrowch}

\begin{document}
\sloppy

\title{TensorKrowch: Smooth integration of tensor networks in machine learning}

\author{Jos\'e Ram\'on Pareja Monturiol}
\affiliation{Departamento de An\'alisis Matem\'atico, Universidad Complutense de Madrid, 28040 Madrid, Spain}
\affiliation{Instituto de Ciencias Matem\'aticas (CSIC-UAM-UC3M-UCM), 28049 Madrid, Spain}

\author{David P\'erez-Garc\'ia}
\affiliation{Departamento de An\'alisis Matem\'atico, Universidad Complutense de Madrid, 28040 Madrid, Spain}
\affiliation{Instituto de Ciencias Matem\'aticas (CSIC-UAM-UC3M-UCM), 28049 Madrid, Spain}

\author{Alejandro Pozas-Kerstjens}
\affiliation{Instituto de Ciencias Matem\'aticas (CSIC-UAM-UC3M-UCM), 28049 Madrid, Spain}

\begin{abstract}
Tensor networks are factorizations of high-dimensional tensors into networks of smaller tensors.
They have applications in physics and mathematics, and recently have been proposed as promising machine learning architectures.
To ease the integration of tensor networks in machine learning pipelines, we introduce \tk{}, an open source Python library built on top of PyTorch.
Providing a user-friendly interface, \tk{} allows users to construct any tensor network, train it, and integrate it as a layer in more intricate deep learning models.
In this paper, we describe the main functionality and basic usage of \tk{}, and provide technical details on its building blocks and the optimizations performed to achieve efficient operation.
\end{abstract}

\maketitle

\section{Introduction}
Tensor networks are factorizations of high-dimensional tensors into network-like structures composed of smaller tensors.
Originating from condensed matter physics and acclaimed for their efficient representation of quantum many-body systems~\cite{fannes1992finitely, white1992density, vidal2003efficient, perez2006matrix, vidal2007entanglement, vidal2008class, evenbly2009algorithms, shi2006classical, tagliacozzo2009simulation, murg2010simulating}, these structures have allowed researchers to comprehend the intricate properties of such systems and, additionally, simulate them using classical computers \cite{hemery2019mps,lin2022simulation,soejima2020isometric}.
Notably, tensor networks are the most successful method for simulating the results of quantum advantage experiments~\cite{pan2022solving,oh2023gbs,rosnet}.
Furthermore, tensor networks were rediscovered within the numerical linear algebra community~\cite{schollwock2005density, oseledets2011tensor, oseledets2011dmrg}, where the techniques have been adapted to other high-dimensional problems such as numerical integration \cite{dolgov2020}, signal processing \cite{sidiropoulos2017}, or epidemic modelling \cite{dolgov2022}.

With the advent of machine learning and the the quest for expressive yet easy-to-train models, tensor networks have been suggested as promising candidates, due to their ability to parameterize regions of the complex space of size exponential in the number of input features.
Since the pioneering works \cite{stoudenmire2016supervised,novikov2016exponential} that used simple, 1-dimensional networks known as Matrix Product States (MPS) in the physics literature \cite{perez2006matrix,cirac2021matrix} and as Tensor Trains in the numerical linear algebra literature \cite{oseledets2011tensor}, these have been applied in both supervised and unsupervised learning settings \cite{glasser2020probabilistic,miller2021tensor,lopezpiqueres2023symmetric}.
Recent studies have also delved into alternative architectures, including Tree Tensor Networks (TTN)~\cite{liu2019machine, cheng2019tree} and Projected Entangled Pair States (PEPS)~\cite{vieijra2022generative, verstraete2006criticality}. 

While there exist indications that tensor network architectures may outperform neural networks in certain scenarios~\cite{wang2020anomaly}, neural networks still hold the upper hand both in versatility and efficiency.
However, there exists a growing number of cases where tensor networks seem to provide advantages. 
First, tensor networks offer a means to compress the matrices used in existing neural networks. 
This process, known as tensorization, reduces the amount of memory required to store the model, and improves the efficiency of the model in both training and inference \cite{novikov2015tensorizing}.
The potential of tensorization has already been explored in several studies~\cite{novikov2015tensorizing, lebedev2014speeding, ma2019tensorized}, offering a way to execute complex models in edge computing devices~\cite{zhang2021ultra}.
Second, the large expertise of the community of quantum many-body physics in tensor networks, and their inspiration in real physical systems, allows to better understand questions related to explainability \cite{liu2019machine,tangpanitanon2022explainable,aizpurua2024explainable}.
Third, this expertise can also bring novel features, such as guarantees on privacy that do not compromise on model performance \cite{pozas2022physics}.
Finally, another promising research line involves the integration of tensor network layers with neural network layers.
For instance, Ref.~\cite{glasser2020probabilistic} proposes using the output of a convolutional neural network, treated as a feature extractor, as the input to four 1-D tensor networks.
Remarkably, this straightforward model achieves near-state-of-the-art performance on the FashionMNIST dataset~\cite{xiao2017fashion}.

Therefore, there are several reasons to believe that the integration of tensor networks into deep learning pipelines can enhance the capabilities of current models. Several libraries exist \cite{miller2019torchmps,usvyatsov2022tntorch,kossaifi2016tensorly,roberts2019tensornetwork,quimb} that allow to use some concrete tensor network architectures as machine learning models, or that use gradient-based methods to optimize tensor-network ansatzes for quantum many-body calculations.
However, from the point of view of machine learning, there is still a need for extensive research to determine, for instance, the situations in which the properties of tensor networks can be maximally leveraged, which are the most effective training methods, optimal architectures (and architectures beyond those used by physicists), and so on.
Consequently, there is a demand for user-friendly tools that enable rapid experimentation in this field.

To address this demand, here we introduce \tk{}\footnote{This work describes the library at its latest version upon publication, namely 1.1.4.}~\cite{pareja2023tensorkrowch}, a Python library built on top of PyTorch~\cite{paszke_pytorch_2019} that aims to bring the full power of tensor networks to machine learning practitioners.
\tk{} allows to construct any tensor network model using the familiar language and capabilities of PyTorch.
The key strength of \tk{} lies in defining a solid set of basic components, namely \texttt{Nodes} and \texttt{Edges}, upon which the entire tensor network can be built.
By connecting these \texttt{Nodes}, a complete \texttt{TensorNetwork} model can be created, that integrates smoothly with other PyTorch modules.
Consequently, \tk{} leverages the full power of PyTorch, including GPU acceleration, automatic differentiation, compilation to XLA, and easy composition of multi-layer networks.
Additionally, \tk{} incorporates built-in implementations of widely used tensor networks such as MPS, MPO, TTN and PEPS, methods to tensorize layers of pre-trained neural network models, and tools developed within the context of quantum information theory that enable interpreting the inner workings of the models \cite{aizpurua2024explainable,tangpanitanon2022explainable,liu2019machine}, such as the computation of reduced density matrices and entanglement entropies.

This work is structured as follows: \Cref{sec:tn} introduces tensor networks and their graphical notation. \Cref{sec:tensorkrowch} presents the library, discussing its underlying philosophy, basic requirements, and main components. \Cref{sec:components} provides a detailed explanation of the components comprising \tk{}, such as \texttt{Nodes}, \texttt{Edges}, and \texttt{TensorNetworks}, and how they are interconnected. \Cref{sec:operations} discusses the operations one can perform between nodes. \Cref{sec:training} combines all the previously described pieces to guide readers in building their own custom models and training them. \Cref{sec:memory} covers advanced concepts like memory management. Lastly, \Cref{sec:additional} contains additional software information such as future development and contribution guidelines, and \Cref{sec:conclusions} presents some concluding remarks.

This paper aims to be self-contained, providing a glimpse of the basics of tensor networks for readers from both the machine learning and quantum physics backgrounds, and introducing the fundamental components of \tk{} in order to enable readers to create and train their own models.
However, it is not intended as a comprehensive tutorial on all the capabilities of the library or a complete description of all its functionality.
Such information is available in the \tk{} documentation at: \url{https://joserapa98.github.io/tensorkrowch}.

\section{Tensor Networks}\label{sec:tn}
In this section, we will introduce the concept of a tensor network and the basic operations that are relevant in the context of machine learning.
For a more in-depth analysis, we refer the reader to \cite{perez2006matrix,orus2014practical,bridgeman2017hand,cirac2021matrix}.

Tensors are an extension of vectors and matrices to higher dimensions. They can be visualized as collections of indexed numbers arranged in multi-dimensional arrays.
In general, a rank-$r$ tensor with dimensions $d_1 \times \cdots \times d_r$ belongs to the tensor product vector space ${\bigotimes_{i=1}^{r}\mathbb{C}^{d_i}\simeq\mathbb{C}^{\times_{i=1}^{r}d_i}}$. Its elements are represented by $T_{i_1\cdots i_r}$, where each $i_j\in [d_j]$.

The literature on tensor networks also presents a useful and practical graphical notation, where tensors are represented as the nodes of a graph with edges corresponding to their indices. In this notation, vectors, matrices and arbitrary tensors take the following form:

\begin{equation*}
    v_i = \begin{tikzpicture}
	\begin{pgfonlayer}{nodelayer}
		\node [style=leaf Node] (0) at (0, 0) {};
		\node [style=none] (1) at (0, 1.25) {};
		\node [style=none] (3) at (0, 1.575) {$_i$};
	\end{pgfonlayer}
	\begin{pgfonlayer}{edgelayer}
		\draw (0) to (1.center);
	\end{pgfonlayer}
\end{tikzpicture}

\qquad
A_{ij} = \begin{tikzpicture}
	\begin{pgfonlayer}{nodelayer}
		\node [style=leaf Node] (0) at (0, 0) {};
		\node [style=none] (1) at (0, 1.25) {};
		\node [style=none] (3) at (0, 1.575) {$_j$};
		\node [style=none] (4) at (0, -1.25) {};
		\node [style=none] (5) at (0, -1.575) {$_i$};
	\end{pgfonlayer}
	\begin{pgfonlayer}{edgelayer}
		\draw (0) to (1.center);
		\draw (0) to (4.center);
	\end{pgfonlayer}
\end{tikzpicture}
\qquad
T_{i_1\cdots i_r} = \begin{tikzpicture}
	\begin{pgfonlayer}{nodelayer}
		\node [style=leaf ParamNode, minimum width=1.2cm] (0) at (0, 0) {};
		\node [style=none] (1) at (-0.8, 1.25) {};
		\node [style=none] (3) at (-0.8, 1.575) {$_{i_1}$};
		\node [style=none] (5) at (0, 1) {$\cdots$};
		\node [style=none] (8) at (0.8, 1.25) {};
		\node [style=none] (9) at (0.8, 1.575) {$_{i_r}$};
		\node [style=none] (10) at (-0.8, 0) {};
		\node [style=none] (11) at (0.8, 0) {};
	\end{pgfonlayer}
	\begin{pgfonlayer}{edgelayer}
		\draw (1.center) to (10.center);
		\draw (8.center) to (11.center);
	\end{pgfonlayer}
\end{tikzpicture}
\end{equation*}

In certain areas where tensors are relevant, for instance in geometry or general relativity, subscripts and superscripts are employed to denote indices in covariant or contravariant spaces, and changing the position of the indices requires using the associated metric.
However, in the tensor network literature it is customary to obviate all these subtleties, and make no distinctions between sub- and superscript indices.
Moreover, it is standard to group and ungroup sets of indices whenever it is convenient.
In this way, the same collection of numbers can be arranged in a matrix $A_{ij}\in\mathbb{C}^{d_1}\otimes \mathbb{C}^{d_2}$, or in a vector $A_k\in\mathbb{C}^{d_1\times d_2}$ by defining $k = (i-1)d_2 + j$.

\begin{equation}
    \begin{tikzpicture}
	\begin{pgfonlayer}{nodelayer}
		\node [style=leaf ParamNode, minimum width=1.2cm] (0) at (0, 0) {};
		\node [style=none] (1) at (-0.8, 1.25) {};
		\node [style=none] (3) at (-0.8, 1.575) {$_{i_1}$};
		\node [style=none] (5) at (0, 1) {$\cdots$};
		\node [style=none] (8) at (0.8, 1.25) {};
		\node [style=none] (9) at (0.8, 1.575) {$_{i_r}$};
		\node [style=none] (10) at (-0.8, 0) {};
		\node [style=none] (11) at (0.8, 0) {};
	\end{pgfonlayer}
	\begin{pgfonlayer}{edgelayer}
		\draw (1.center) to (10.center);
		\draw (8.center) to (11.center);
	\end{pgfonlayer}
\end{tikzpicture} = \begin{tikzpicture}
	\begin{pgfonlayer}{nodelayer}
		\node [style=leaf ParamNode, minimum width=1.2cm] (0) at (0, 0) {};
		\node [style=none] (1) at (-0.8, 1.25) {};
		\node [style=none] (3) at (-0.8, 1.575) {$_{i_1}$};
		\node [style=none] (5) at (0, 1) {$\cdots$};
		\node [style=none] (8) at (0.8, 1.25) {};
		\node [style=none] (9) at (0.8, 1.575) {$_{i_k}$};
		\node [style=none] (10) at (-0.8, 0) {};
		\node [style=none] (11) at (0.8, 0) {};
		\node [style=none] (12) at (-0.8, -1.275) {};
		\node [style=none] (13) at (-0.8, -1.6) {$_{i_{k+1}}$};
		\node [style=none] (14) at (0, -1.025) {$\cdots$};
		\node [style=none] (15) at (0.8, -1.275) {};
		\node [style=none] (16) at (0.8, -1.6) {$_{i_r}$};
		\node [style=none] (17) at (-0.8, -0.025) {};
		\node [style=none] (18) at (0.8, -0.025) {};
	\end{pgfonlayer}
	\begin{pgfonlayer}{edgelayer}
		\draw (1.center) to (10.center);
		\draw (8.center) to (11.center);
		\draw (12.center) to (17.center);
		\draw (15.center) to (18.center);
	\end{pgfonlayer}
\end{tikzpicture} \simeq \begin{tikzpicture}
	\begin{pgfonlayer}{nodelayer}
		\node [style=leaf Node] (0) at (0, 0) {};
		\node [style=none] (1) at (0, 1.25) {};
		\node [style=none] (3) at (0, 1.575) {$_\alpha$};
		\node [style=none] (12) at (0, -1.275) {};
		\node [style=none] (13) at (0, -1.6) {$_\beta$};
	\end{pgfonlayer}
	\begin{pgfonlayer}{edgelayer}
		\draw [ultra thick] (0) to (1.center);
		\draw [ultra thick] (0) to (12.center);
	\end{pgfonlayer}
\end{tikzpicture}
\end{equation}
This flexibility in representation makes it clear that tensors are essentially linear mappings between tensor product spaces, similar to how matrices represent linear mappings between vector spaces.
Furthermore, it provides a convenient framework to decompose tensors via matrix factorization algorithms, as we will see shortly below.

Indices of tensors can be \textit{contracted}.
The contraction of indices (of the same or different tensors) consists of a generalization of the scalar product between vectors, this is, the sum of products of the elements in the corresponding axes.
For two arbitrary tensors, $R$ and $S$, of ranks $r$ and $s$, respectively, the contraction over one of its indices is given by

\begin{equation}
    \sum_{\alpha=1}^{d}{R_{i_1\ldots i_{m-1}\alpha i_{m+1}\ldots i_r}S_{j_1\ldots j_{n-1}\alpha j_{n+1}\ldots j_s}} = T_{i_1\ldots i_{m-1} i_{m+1}\ldots i_r}^{j_1\ldots j_{n-1} j_{n+1}\ldots j_s},
\end{equation}
where we have assumed that the dimensions of the indices $i_m$ and $j_n$ are both $d$.
This tensor $T$ is of rank $(r+s-2)$.
In the graphical notation, the contraction is represented by connecting the corresponding edges of the tensors:

\begin{equation}
    \begin{tikzpicture}
	\begin{pgfonlayer}{nodelayer}
		\node [style=leaf Node] (0) at (-1.25, 0) {};
		\node [style=none] (1) at (-2.5, 1) {};
		\node [style=none] (12) at (-2.5, -1) {};
		\node [style=leaf Node] (13) at (1.25, 0) {};
		\node [style=none] (14) at (2.5, 1) {};
		\node [style=none] (15) at (2.5, -1) {};
		\node [style=none] (16) at (0, 0.325) {$_\alpha$};
		\node [style=none] (17) at (-2.5, 0.2) {$\vdots$};
		\node [style=none] (18) at (2.5, 0.2) {$\vdots$};
		\node [style=none] (19) at (-2.825, 1.25) {$_{i_1}$};
		\node [style=none] (20) at (-2.825, -1.25) {$_{i_r}$};
		\node [style=none] (21) at (3, 1.25) {$_{j_1}$};
		\node [style=none] (22) at (3, -1.25) {$_{j_s}$};
	\end{pgfonlayer}
	\begin{pgfonlayer}{edgelayer}
		\draw (0) to (1.center);
		\draw (0) to (12.center);
		\draw (0) to (13);
		\draw (14.center) to (13);
		\draw (13) to (15.center);
	\end{pgfonlayer}
\end{tikzpicture} = \begin{tikzpicture}
	\begin{pgfonlayer}{nodelayer}
		\node [style=none] (1) at (-1.75, 1) {};
		\node [style=none] (12) at (-1.75, -1) {};
		\node [style=none] (14) at (1.75, 1) {};
		\node [style=none] (15) at (1.75, -1) {};
		\node [style=none] (17) at (-1.5, 0.2) {$\vdots$};
		\node [style=none] (18) at (1.5, 0.2) {$\vdots$};
		\node [style=none] (19) at (-2.075, 1.25) {$_{i_1}$};
		\node [style=none] (20) at (-2.075, -1.25) {$_{i_r}$};
		\node [style=none] (21) at (2.25, 1.25) {$_{j_1}$};
		\node [style=none] (22) at (2.25, -1.25) {$_{j_s}$};
		\node [style=none] (23) at (-0.5, 1) {};
		\node [style=none] (24) at (-0.5, -1) {};
		\node [style=none] (25) at (0.5, 1) {};
		\node [style=none] (26) at (0.5, -1) {};
		\node [style=leaf ParamNode, minimum height=1.2cm] (27) at (0, 0) {};
	\end{pgfonlayer}
	\begin{pgfonlayer}{edgelayer}
		\draw (12.center) to (24.center);
		\draw (1.center) to (23.center);
		\draw (26.center) to (15.center);
		\draw (25.center) to (14.center);
	\end{pgfonlayer}
\end{tikzpicture}
\end{equation}

Contraction enables connecting nodes to form graphs, with the only requirement being that the nodes must have the same dimensions along the connected axes.
By contracting all the connected edges in the graph, one can then contract a whole tensor network to obtain a single tensor that preserves the remaining dangling edges:

\begin{equation}
    \begin{tikzpicture}
	\begin{pgfonlayer}{nodelayer}
		\node [style=leaf Node] (0) at (-2, 2) {};
		\node [style=leaf Node] (1) at (-1, -0.25) {};
		\node [style=leaf Node] (2) at (0, 1.5) {};
		\node [style=leaf Node] (3) at (2, -0.75) {};
		\node [style=leaf Node] (4) at (0.25, -2) {};
		\node [style=none] (5) at (-3.5, 2) {};
		\node [style=none] (6) at (-2.5, -0.25) {};
		\node [style=none] (7) at (3.5, -0.75) {};
	\end{pgfonlayer}
	\begin{pgfonlayer}{edgelayer}
		\draw (0) to (1);
		\draw (2) to (1);
		\draw (3) to (2);
		\draw (4) to (3);
		\draw (1) to (3);
		\draw [red] (5.center) to (0);
		\draw [red] (6.center) to (1);
		\draw [red] (3) to (7.center);
	\end{pgfonlayer}
\end{tikzpicture} = \begin{tikzpicture}
	\begin{pgfonlayer}{nodelayer}
		\node [style=leaf Node] (0) at (0, 0) {};
		\node [style=none] (1) at (-1.25, 0.75) {};
		\node [style=none] (2) at (-1.25, -0.5) {};
		\node [style=none] (3) at (1.25, 0) {};
	\end{pgfonlayer}
	\begin{pgfonlayer}{edgelayer}
		\draw [red] (1.center) to (0);
		\draw [red] (2.center) to (0);
		\draw [red] (0) to (3.center);
	\end{pgfonlayer}
\end{tikzpicture}
\end{equation}

Conversely, one may obtain a tensor network by splitting tensors via the various existing decomposition methods~\cite{kolda2009tensor}.
A commonplace method is exploiting grouping and ungrouping of indices to reshape tensors as matrices and applying the singular value decomposition, 

\begin{equation}
    \begin{tikzpicture}
	\begin{pgfonlayer}{nodelayer}
		\node [style=none] (1) at (-1.75, 1) {};
		\node [style=none] (12) at (-1.75, -1) {};
		\node [style=none] (14) at (1.75, 1) {};
		\node [style=none] (15) at (1.75, -1) {};
		\node [style=none] (17) at (-1.5, 0.2) {$\vdots$};
		\node [style=none] (18) at (1.5, 0.2) {$\vdots$};
		\node [style=none] (19) at (-2.075, 1.25) {$_{i_1}$};
		\node [style=none] (20) at (-2.075, -1.25) {$_{i_r}$};
		\node [style=none] (21) at (2.25, 1.25) {$_{j_1}$};
		\node [style=none] (22) at (2.25, -1.25) {$_{j_s}$};
		\node [style=none] (23) at (-0.5, 1) {};
		\node [style=none] (24) at (-0.5, -1) {};
		\node [style=none] (25) at (0.5, 1) {};
		\node [style=none] (26) at (0.5, -1) {};
		\node [style=leaf ParamNode, minimum height=1.2cm] (27) at (0, 0) {};
	\end{pgfonlayer}
	\begin{pgfonlayer}{edgelayer}
		\draw (12.center) to (24.center);
		\draw (1.center) to (23.center);
		\draw (26.center) to (15.center);
		\draw (25.center) to (14.center);
	\end{pgfonlayer}
\end{tikzpicture} \simeq \begin{tikzpicture}
	\begin{pgfonlayer}{nodelayer}
		\node [style=leaf Node] (0) at (0, 0) {};
		\node [style=none] (1) at (1.5, 0) {};
		\node [style=none] (12) at (-1.5, 0) {};
		\node [style=none] (19) at (-1.825, 0) {$_\alpha$};
		\node [style=none] (20) at (1.825, 0) {$_\beta$};
		\node [style=none] (21) at (0, 1) {};
		\node [style=none] (22) at (0, -1) {};
	\end{pgfonlayer}
	\begin{pgfonlayer}{edgelayer}
		\draw [ultra thick] (0) to (1.center);
		\draw [ultra thick] (0) to (12.center);
	\end{pgfonlayer}
        \draw [style={dashed_edge}, red] (21.center) to (22.center);
\end{tikzpicture} = \begin{tikzpicture}
	\begin{pgfonlayer}{nodelayer}
		\node [style=data Node] (0) at (-1.75, 0) {};
		\node [style=none] (1) at (3.25, 0) {};
		\node [style=none] (12) at (-3.25, 0) {};
		\node [style=none] (19) at (-3.575, 0) {$_\alpha$};
		\node [style=none] (20) at (3.575, 0) {$_\beta$};
		\node [style=data Node] (21) at (1.75, 0) {};
		\node [style=leaf ParamNode, fill={rgb,255: red,132; green,220; blue,198}, minimum width=0.3cm, minimum height=0.3cm, rotate=45] (22) at (0, 0) {};
	\end{pgfonlayer}
	\begin{pgfonlayer}{edgelayer}
		\draw [ultra thick] (0) to (12.center);
		\draw [ultra thick] (21) to (1.center);
		\draw (0) to (21);
	\end{pgfonlayer}
\end{tikzpicture} \simeq \begin{tikzpicture}
	\begin{pgfonlayer}{nodelayer}
		\node [style=data Node] (0) at (-1.75, 0) {};
		\node [style=data Node] (21) at (1.75, 0) {};
		\node [style=leaf ParamNode, fill={rgb,255: red,132; green,220; blue,198}, minimum width=0.3cm, minimum height=0.3cm, rotate=45] (22) at (0, 0) {};
		\node [style=none] (23) at (-3, 1) {};
		\node [style=none] (24) at (-3, -1) {};
		\node [style=none] (25) at (-3, 0.2) {$\vdots$};
		\node [style=none] (26) at (-3.325, 1.25) {$_{i_1}$};
		\node [style=none] (27) at (-3.325, -1.25) {$_{i_r}$};
		\node [style=none] (28) at (3, 1) {};
		\node [style=none] (29) at (3, -1) {};
		\node [style=none] (30) at (3, 0.2) {$\vdots$};
		\node [style=none] (31) at (3.5, 1.25) {$_{j_1}$};
		\node [style=none] (32) at (3.5, -1.25) {$_{j_s}$};
	\end{pgfonlayer}
	\begin{pgfonlayer}{edgelayer}
		\draw (0) to (21);
		\draw (23.center) to (0);
		\draw (24.center) to (0);
		\draw (21) to (28.center);
		\draw (21) to (29.center);
	\end{pgfonlayer}
\end{tikzpicture},
\end{equation}
since this provides, in many cases of relevance, a form of the tensor with desirable properties \cite{perez2006matrix}.
In particular, truncations of the singular value decomposition give the best low-rank approximations of the original tensor under the Frobenius and $\ell_2$ norms.

In machine learning, there are two main applications of tensor networks.
The first is the decomposition, or approximation, of large tensors in networks with a smaller number of entries.
This procedure is known as tensorization~\cite{novikov2015tensorizing, lebedev2014speeding, ma2019tensorized}.
Tensorization allows not only to reduce the number of elements to be stored (and thus reducing the storage memory), but also to reduce the computation time.
Consider a linear layer within a neural network, where both the input and output dimensions are $d^n$.
The multiplication of the corresponding weight matrix with the input data vector requires $O(d^{2n})$ operations.

On the other hand, consider the same operation between factorizations (or approximations) of the vector and the matrix into tensor networks.
For this example, let us use the well-known Matrix Product State (MPS)~\cite{perez2006matrix, cirac2021matrix} form for the vector and a Matrix Product Operator (MPO)~\cite{pirvu2010matrix} form for the matrix, both with $n$ nodes, each equipped with dangling edges of dimension $d$ (this is known as the \textit{physical dimension} in the physics literature) and connected to their neighbors via edges of dimension $D$ (this is known as the \textit{bond dimension}).
Thus, the MPS requires only $O(ndD^2)$ elements to represent the vector, and the MPO
requires $O(nd^2D^2)$ elements to represent the matrix.

\begin{equation}
    \begin{tikzpicture}
	\begin{pgfonlayer}{nodelayer}
		\node [style=data Node] (0) at (-4, 0) {};
		\node [style=none] (1) at (-4, 1.25) {};
		\node [style=none] (3) at (-4.325, 1.075) {$_d$};
		\node [style=data Node] (4) at (-2, 0) {};
		\node [style=none] (5) at (-2, 1.25) {};
		\node [style=data Node] (7) at (0, 0) {};
		\node [style=none] (8) at (0, 1.25) {};
		\node [style=data Node] (10) at (2, 0) {};
		\node [style=none] (11) at (2, 1.25) {};
		\node [style=data Node] (13) at (4, 0) {};
		\node [style=none] (14) at (4, 1.25) {};
		\node [style=none] (16) at (-3, 0.375) {$_D$};
		\node [style=none] (17) at (0, 2) {MPS};
	\end{pgfonlayer}
	\begin{pgfonlayer}{edgelayer}
		\draw (0) to (1.center);
		\draw (4) to (5.center);
		\draw (7) to (8.center);
		\draw (10) to (11.center);
		\draw (13) to (14.center);
		\draw (0) to (4);
		\draw (4) to (7);
		\draw (7) to (10);
		\draw (10) to (13);
	\end{pgfonlayer}
\end{tikzpicture} \qquad \begin{tikzpicture}
	\begin{pgfonlayer}{nodelayer}
		\node [style=leaf Node] (0) at (-4, 0) {};
		\node [style=none] (1) at (-4, 1.25) {};
		\node [style=none] (3) at (-4.325, 1.075) {$_d$};
		\node [style=leaf Node] (4) at (-2, 0) {};
		\node [style=none] (5) at (-2, 1.25) {};
		\node [style=leaf Node] (7) at (0, 0) {};
		\node [style=none] (8) at (0, 1.25) {};
		\node [style=leaf Node] (10) at (2, 0) {};
		\node [style=none] (11) at (2, 1.25) {};
		\node [style=leaf Node] (13) at (4, 0) {};
		\node [style=none] (14) at (4, 1.25) {};
		\node [style=none] (16) at (-3, 0.375) {$_D$};
		\node [style=none] (17) at (-4, -1.25) {};
		\node [style=none] (18) at (-2, -1.25) {};
		\node [style=none] (19) at (0, -1.25) {};
		\node [style=none] (20) at (2, -1.25) {};
		\node [style=none] (21) at (4, -1.25) {};
		\node [style=none] (22) at (-4.325, -0.925) {$_d$};
		\node [style=none] (23) at (0, 2) {MPO};
		\node [style=none] (24) at (5.25, 1.5) {Output};
		\node [style=none] (25) at (5.25, -1.25) {Input};
	\end{pgfonlayer}
	\begin{pgfonlayer}{edgelayer}
		\draw (0) to (1.center);
		\draw (4) to (5.center);
		\draw (7) to (8.center);
		\draw (10) to (11.center);
		\draw (13) to (14.center);
		\draw (0) to (4);
		\draw (4) to (7);
		\draw (7) to (10);
		\draw (10) to (13);
		\draw (0) to (17.center);
		\draw (4) to (18.center);
		\draw (7) to (19.center);
		\draw (10) to (20.center);
		\draw (13) to (21.center);
	\end{pgfonlayer}
\end{tikzpicture}
\end{equation}

The vector-matrix multiplication is carried out by contracting the MPS with the MPO, connecting all the dangling edges of the MPS to the input edges of the MPO:

\begin{equation}
    \begin{tikzpicture}
	\begin{pgfonlayer}{nodelayer}
		\node [style=leaf Node] (0) at (-4, 0.75) {};
		\node [style=none] (1) at (-4, 2) {};
		\node [style=leaf Node] (4) at (-2, 0.75) {};
		\node [style=none] (5) at (-2, 2) {};
		\node [style=leaf Node] (7) at (0, 0.75) {};
		\node [style=none] (8) at (0, 2) {};
		\node [style=leaf Node] (10) at (2, 0.75) {};
		\node [style=none] (11) at (2, 2) {};
		\node [style=leaf Node] (13) at (4, 0.75) {};
		\node [style=none] (14) at (4, 2) {};
		\node [style=data Node] (24) at (-4, -1.25) {};
		\node [style=data Node] (27) at (-2, -1.25) {};
		\node [style=data Node] (29) at (0, -1.25) {};
		\node [style=data Node] (31) at (2, -1.25) {};
		\node [style=data Node] (33) at (4, -1.25) {};
	\end{pgfonlayer}
	\begin{pgfonlayer}{edgelayer}
 \node [style=virtual ParamNode, minimum width=0.9cm, minimum height=1.8cm] (34) at (-4, -0.25) {};
		\node [style=virtual ParamNode, minimum width=0.9cm, minimum height=1.8cm] (35) at (-2, -0.25) {};
		\node [style=virtual ParamNode, minimum width=0.9cm, minimum height=1.8cm] (36) at (0, -0.25) {};
		\node [style=virtual ParamNode, minimum width=0.9cm, minimum height=1.8cm] (37) at (2, -0.25) {};
		\node [style=virtual ParamNode, minimum width=0.9cm, minimum height=1.8cm] (38) at (4, -0.25) {};
		\draw (0) to (1.center);
		\draw (4) to (5.center);
		\draw (7) to (8.center);
		\draw (10) to (11.center);
		\draw (13) to (14.center);
		\draw (0) to (4);
		\draw (4) to (7);
		\draw (7) to (10);
		\draw (10) to (13);
		\draw (24) to (27);
		\draw (27) to (29);
		\draw (29) to (31);
		\draw (31) to (33);
		\draw (0) to (24);
		\draw (4) to (27);
		\draw (7) to (29);
		\draw (10) to (31);
		\draw (13) to (33);
	\end{pgfonlayer}
\end{tikzpicture} = \begin{tikzpicture}
	\begin{pgfonlayer}{nodelayer}
		\node [style=resultant Node] (0) at (-4, 0) {};
		\node [style=none] (1) at (-4, 1.25) {};
		\node [style=none] (3) at (-4.325, 1.075) {$_d$};
		\node [style=resultant Node] (4) at (-2, 0) {};
		\node [style=none] (5) at (-2, 1.25) {};
		\node [style=resultant Node] (7) at (0, 0) {};
		\node [style=none] (8) at (0, 1.25) {};
		\node [style=resultant Node] (10) at (2, 0) {};
		\node [style=none] (11) at (2, 1.25) {};
		\node [style=resultant Node] (13) at (4, 0) {};
		\node [style=none] (14) at (4, 1.25) {};
		\node [style=none] (16) at (-3, 0.375) {$_{D^2}$};
	\end{pgfonlayer}
	\begin{pgfonlayer}{edgelayer}
		\draw (0) to (1.center);
		\draw (4) to (5.center);
		\draw (7) to (8.center);
		\draw (10) to (11.center);
		\draw (13) to (14.center);
		\draw (0) to (4);
		\draw (4) to (7);
		\draw (7) to (10);
		\draw (10) to (13);
	\end{pgfonlayer}
\end{tikzpicture}
    \label{eq:mpsmpo}
\end{equation}

The contraction of these connected edges requires only $O(nd^2D^4)$ operations.
Note that the contractions in the boxes in Eq.~\eqref{eq:mpsmpo} all involve different tensors and thus can be performed in parallel, thereby giving even greater savings in practice.
Thus, representing linear layers in tensorized form presents advantages in terms of, both, memory usage and in the time taken to perform the operations. Similar advantages are also present in the backward pass of automatic differentiation routines \cite{novikov2015tensorizing}.

The second main application of tensor networks in machine learning is as architectures themselves.
In tensor network architectures, every cell of every tensor is a trainable parameter, and so they implement linear models operating in high-dimensional tensor-product spaces.
Taking the example of classification, the contraction of a tensor form of the input with the tensor network gives the corresponding prediction, that is used to compute a loss function whose gradients are used to adjust the model parameters.
This approach was pioneered by \cite{stoudenmire2016supervised,novikov2016exponential} with MPS, and since has been applied to many different problems~\cite{wang2020anomaly,martyn2020entanglement,miller2021tensor} and network architectures~\cite{liu2019machine,cheng2019tree,vieijra2022generative,reyes2021multi}.

\section{\tk{}}\label{sec:tensorkrowch}
\subsection{Motivation}
The works \cite{stoudenmire2016supervised,novikov2016exponential} gave rise to the area of tensor network machine learning, where nowadays many different network architectures are being used for different purposes.
However, the common choices for architectures are inherited from the physics community, which has broad experience in tensor networks but is restricted to 1-D and select \mbox{2-D} architectures.
It is for this reason that, currently, there are libraries that are well-integrated into deep learning frameworks but are optimized for specific tensor networks \cite{miller2019torchmps} or put their focus on tensorization \cite{usvyatsov2022tntorch,kossaifi2016tensorly}.
On the other hand, there exist libraries like TensorNetwork~\cite{roberts2019tensornetwork}, that offer more generality in terms of architectures at the cost of more complications at the time of integrating them into machine learning pipelines.
Furthermore, the available libraries tend to have a broader focus on physics applications, and thus features that are of interest in machine learning, such as custom parameter initializations, hyperparameter optimization, or even the construction of complicated models, are not considered.

\tk{} is developed with the aim of providing a comprehensive framework where one can rapidly prototype tensor network layers as standalone models and integrate them in deep machine learning models.
Its main characteristics are the following:\\[5pt]
\textbf{Generality:} \tk{} allows users to construct any tensor network by creating and connecting nodes.
Users can selectively choose which nodes to train and to define arbitrary operations between their components.
In addition to this, \tk{} has pre-built classes for the most common types of networks.\\[5pt]
\textbf{Ease of use:} At the core of \tk{} is a Pythonic approach, presenting a simple interface with building blocks and operations that are combined in order to create complex models and training pipelines.
These primary objects and operations are described in sections \ref{sec:components} and \ref{sec:operations}, respectively.\\[5pt]
\textbf{Optimization:} While the interface remains simple, numerous optimizations are implemented in the background in order to perform operations efficiently and to reduce redundant computations during training.
\Cref{sec:memory} details the optimizations that are performed.\\[5pt]
\textbf{Integration:} \tk{} is written on top of a well-established deep learning framework, namely PyTorch \cite{paszke_pytorch_2019}.
This integration enables the creation of \texttt{TensorNetwork} models that function like any other PyTorch layer, seamlessly combining with existing components.
Consequently, \tk{} fully leverages the capabilities of PyTorch, including GPU acceleration, automatic differentiation, and easy composition of multi-layer networks.\\[5pt]
Moreover, \tk{} can be also used for tensorization purposes, by substituting or approximating dense matrices in deep PyTorch models and applying the built-in matrix factorization techniques.

\subsection{Installation and requirements}
\tk{} is a Python library available on Linux, Mac and Windows operating systems. It can be installed via \texttt{pip} with the following command line:

\noindent\begin{minipage}[c]{0.31\textwidth}
\begin{lstlisting}
 pip install tensorkrowch
\end{lstlisting}
\end{minipage}

The basic requirement of \tk{} is PyTorch~\cite{paszke_pytorch_2019}, which is used as machine learning backend.
Additionally, \tk{} requires opt\_einsum~\cite{daniel2018opt}, used in the implementation of the \texttt{einsum} operation and that allows for the automatic search of good network contraction paths via greedy algorithms.

The source code for \tk{} is hosted on GitHub at

\centerline{\url{https://github.com/joserapa98/tensorkrowch}}

\noindent and is distributed under the MIT License.
More details about the software, packaging information, and guidelines for contributing to \tk{} are included in Sec. \ref{sec:additional}.

\subsection{Basic Usage}\label{sec:basic}
\tk{} provides a set of basic components, namely \texttt{Node}, \texttt{Edge}, \texttt{TensorNetwork}, and variants of these, that can be combined to build trainable models.
The usual workflow consists in the following steps:
1) Define the structure of the graph by creating nodes and connecting them; and initialize the tensors within those.
2) Specify which nodes will be used to store the input tensors coming from the training dataset.
3) Define the contraction algorithm to reduce the whole network to a single output tensor, which one can input to any other layer that might follow the tensor network.

To carry out these steps, one needs to know how to create and combine the different building blocks of the model, and how to operate them to contract the network. These topics will be covered in detail in \Cref{sec:components} and \Cref{sec:operations}, respectively.

Once the custom tensor network is defined, the process of training it is analogous to what one would do in vanilla PyTorch.
To illustrate this with an example, let us introduce one of the built-in classes provided by TensorKrowch, \texttt{MPSLayer}.
This class is a variation of the traditional MPS with an additional node that has a dangling edge representing an output dimension. As a result, the \texttt{MPSLayer} is contracted into a vector of the specified dimension.

\noindent\begin{minipage}[c]{0.60\textwidth}
\begin{lstlisting}
 import torch
 import tensorkrowch as tk

 # Instantiate model
 mps = tk.models.MPSLayer(n_features=1001,
                          in_dim=2,
                          out_dim=10,
                          bond_dim=10)
\end{lstlisting}
\end{minipage}

Tensor network models expect as input a rank-3 tensor with dimensions $b \times n \times d$, being $b$ the size of the batch, $n$ the number of feature vectors, and $d$ their dimension.
This tensor, which is a batch of sequences of vectors, represents a tensor network itself, given by the tensor product of all the vectors corresponding to each batch element.
These vectors will be placed in specific nodes so that the network can be contracted with these new data.

However, data tensors tend to come in the form of $b \times n$ matrices, requiring a previous transformation to get the proper rank-3 tensor.
That is, for each batch element there is a vector of features, that has to be turned into a sequence of feature vectors.
To accomplish this, each feature has to be embedded into a vector space. \tk{} provides five common embedding functions, namely \texttt{unit}, \texttt{add\_ones}, \texttt{poly}, \texttt{discretize} and \texttt{basis}, the first two being introduced in~\cite{stoudenmire2016supervised} and~\cite{novikov2016exponential}, respectively.

\noindent\begin{minipage}[c]{0.60\textwidth}
\begin{lstlisting}
 # Shape: batch_size x n_features
 data = torch.randn(100, 1000) 
 # Shape: batch_size x n_features x in_dim=2
 data = tk.embeddings.unit(data)

 # Shape: batch_size x out_dim=10
 labels = torch.randn(100, 10)
\end{lstlisting}
\end{minipage}

To reduce repetition of ancillary computations due to dealing with the tensor network structure, \tk{} introduces some extra steps that need to be carried out before starting training.
These involve setting memory modes and tracing the model; both advanced features will be covered in \Cref{sec:memory}.

\noindent\begin{minipage}[c]{0.60\textwidth}
\begin{lstlisting}
 # Set memory modes
 mps.auto_stack  = True
 mps.auto_unbind = False

 # Trace
 example = torch.zeros(1, 1000, 2) 
 # Shape: batch_size=1 x n_features x in_dim=2
 mps.trace(example)
\end{lstlisting}
\end{minipage}

To be able to train, one needs to set a loss function to minimize and an optimizer to update the parameters of the model according to their gradients.
It is important that the optimizer is set after the model has been traced, since the parameters of the model might change during this process.

\noindent\begin{minipage}[c]{0.60\textwidth}
\begin{lstlisting}
 loss_fun  = torch.nn.CrossEntropyLoss()
 optimizer = torch.optim.Adam(mps.parameters(),
                              lr=1e-4,
                              weight_decay=1e-2)
\end{lstlisting}
\end{minipage}

Finally, the above ingredients can be put together in the training loop.

\noindent\begin{minipage}[c]{0.60\textwidth}
\begin{lstlisting}
 for epoch in range(n_epochs):
     # Contract tensor network with input data
     scores = mps(data)
     
     # Compute the loss
     loss = loss_fun(scores, labels)
     
     # Backpropagate and update parameters
     optimizer.zero_grad()
     loss.backward()
     optimizer.step()
\end{lstlisting}
\end{minipage}

\section{The Building Blocks}\label{sec:components}
The main structure of \tk{} is similar to that of TensorNetwork \cite{roberts2019tensornetwork}.
Namely, the main object in \tk{} is a \texttt{TensorNetwork}, that is populated with \texttt{Nodes}.
These have \texttt{Edges} that are connected according to the desired structure.
Creating a \texttt{TensorNetwork} is done as follows:

\noindent\begin{minipage}[c]{0.32\textwidth}
\begin{lstlisting}
 import torch
 import tensorkrowch as tk
 net = tk.TensorNetwork()
\end{lstlisting}
\end{minipage}

\texttt{Nodes} are the basic elements that make up a \texttt{TensorNetwork}.
They serve as containers for PyTorch's \texttt{torch.Tensor} objects and hold essential information for building, operating, and training the network.
Key aspects associated with nodes include shape, tensor, axes, edges, network membership, and successors.
Tensors themselves are not actually contained within nodes; instead, they are stored in the shared memory system that is accessible to all nodes in the \texttt{TensorNetwork} (see \Cref{sec:memory} for details).
To create a \texttt{Node} one can specify its name, shape, names for its axes, network, and an initialization method to create a new tensor for it.
Together with the code we present a graphical notation to depict not only the tensor network but all the elements that form the \texttt{TensorNetwork} object.

\noindent\begin{minipage}[c]{0.60\textwidth}
\begin{lstlisting}
 node1 = tk.Node(name="node1",
                 shape=(2, 5, 2),
                 axes_names=
                     ("left", "input", "right"),
                 network=net,
                 init_method="randn")
\end{lstlisting}
\end{minipage}
\hfill
\begin{minipage}[c]{0.39\textwidth}
    \flushright
    \begin{tikzpicture}
	\begin{pgfonlayer}{nodelayer}
		\node [style=leaf Node] (0) at (0, 0) {};
		\node [style=none] (1) at (1.5, 0) {};
		\node [style=none] (2) at (-1.5, 0) {};
		\node [style=none] (3) at (0, -1.5) {};
		\node [style=none] (4) at (-2.05, 0) {\scriptsize{\texttt{left}}};
		\node [style=none] (5) at (2.25, 0) {\scriptsize{\texttt{right}}};
		\node [style=none] (6) at (0, -1.75) {\scriptsize{\texttt{input}}};
		\node [style=none] (7) at (-2.75, 1.15) {};
		\node [style=none] (8) at (3, 1.15) {};
		\node [style=none] (9) at (-2.75, -2.25) {};
		\node [style=none] (10) at (3, -2.25) {};
		\node [style=none] (11) at (-2.25, 1.5) {\texttt{net}};
		\node [style=none] (12) at (0, 0.825) {\scriptsize{\texttt{node1}}};
	\end{pgfonlayer}
	\begin{pgfonlayer}{edgelayer}
		\draw (2.center) to (0);
		\draw (0) to (1.center);
		\draw (0) to (3.center);
		\draw [style={dashed_edge}] (7.center) to (9.center);
		\draw [style={dashed_edge}] (7.center) to (8.center);
		\draw [style={dashed_edge}] (8.center) to (10.center);
		\draw [style={dashed_edge}] (10.center) to (9.center);
	\end{pgfonlayer}
\end{tikzpicture}
\end{minipage}

Specifying a shape automatically creates a set of edges for the \texttt{Node}.
An \texttt{Edge} is nothing more than an object that wraps references to the nodes it connects.
Thus it stores information like the nodes it connects, the corresponding nodes’ axes it is attached to, its size, etc.

To connect nodes, one can access the desired edges using their names, and connect them with the caret operator:

\noindent\begin{minipage}[c]{0.60\textwidth}
\begin{lstlisting}
 # Equivalent to initialize Node with
 # "randn" init_method
 node2 = tk.randn(name="node2",
                  shape=(2, 5, 2),
                  axes_names=
                      ("left", "input", "right"),
                  network=net)

 node1["right"] ^ node2["left"]
\end{lstlisting}
\end{minipage}
\hfill
\begin{minipage}[c]{0.39\textwidth}
    \flushright
    \begin{tikzpicture}
	\begin{pgfonlayer}{nodelayer}
		\node [style=leaf Node] (0) at (-1, 0) {};
		\node [style=leaf Node] (1) at (1, 0) {};
		\node [style=none] (2) at (-2.5, 0) {};
		\node [style=none] (3) at (-1, -1.5) {};
		\node [style=none] (4) at (-3.05, 0) {\scriptsize{\texttt{left}}};
		\node [style=none] (6) at (-1, -1.75) {\scriptsize{\texttt{input}}};
		\node [style=none] (7) at (-3.75, 1.15) {};
		\node [style=none] (8) at (4, 1.15) {};
		\node [style=none] (9) at (-3.75, -2.25) {};
		\node [style=none] (10) at (4, -2.25) {};
		\node [style=none] (11) at (-3.25, 1.5) {\texttt{net}};
		\node [style=none] (12) at (-1, 0.825) {\scriptsize{\texttt{node1}}};
		\node [style=none] (13) at (2.5, 0) {};
		\node [style=none] (14) at (1, -1.5) {};
		\node [style=none] (15) at (1, -1.75) {\scriptsize{\texttt{input}}};
		\node [style=none] (16) at (3.25, 0) {\scriptsize{\texttt{right}}};
		\node [style=none] (17) at (1, 0.825) {\scriptsize{\texttt{node2}}};
	\end{pgfonlayer}
	\begin{pgfonlayer}{edgelayer}
		\draw (2.center) to (0);
		\draw (0) to (1);
		\draw (0) to (3.center);
		\draw [style={dashed_edge}] (7.center) to (9.center);
		\draw [style={dashed_edge}] (7.center) to (8.center);
		\draw [style={dashed_edge}] (8.center) to (10.center);
		\draw [style={dashed_edge}] (10.center) to (9.center);
		\draw [style=Edge] (1) to (13.center);
		\draw [style=Edge] (1) to (14.center);
	\end{pgfonlayer}
\end{tikzpicture}
\end{minipage}

It is important to note that the operation \texttt{\^} does not perform contractions, since it may be done on edges of empty nodes.
As it will be shown below, contraction between tensors is performed by using the \texttt{@} operator, that contracts tensors along all the edges that connect both nodes.

Edges can be designated as \textit{batch} edges if one includes the string \texttt{"batch"} in the name of the corresponding axis.
This allows for performing batch contractions if the nodes involved in the operation both have batch edges sharing the same name.
This functionality is analogous to that of PyTorch functions like, for instance, \texttt{torch.bmm}, used to perform batch matrix multiplication.

\subsection{Types of nodes}
\tk{} features several types of nodes, that have different functionalities and goals within the library.\\[5pt]
\textbf{Parameter nodes:}
In analogy with PyTorch having \texttt{torch.Tensor} to denote a tensor of non-trainable entries and \texttt{torch.nn.Parameter} for tensors of trainable parameters, \tk{} has \texttt{Node} to denote containers of tensors with non-trainable entries and \texttt{ParamNode} to denote containers of tensors with trainable entries.
One can instantiate a \texttt{ParamNode} with:

\noindent\begin{minipage}[c]{0.60\textwidth}
\begin{lstlisting}
 paramnode = tk.ParamNode(name="paramnode",
                          shape=(2, 2),
                          axes_names=
                              ("left", "right"),
                          network=net,
                          init_method="randn")

 paramnode["left"] ^ node1["left"]
 paramnode["right"] ^ node2["right"]
\end{lstlisting}
\end{minipage}
\hfill
\begin{minipage}[c]{0.39\textwidth}
    \flushright
    \begin{tikzpicture}
	\begin{pgfonlayer}{nodelayer}
		\node [style=leaf Node] (0) at (-1, 0) {};
		\node [style=none] (3) at (-1, -1.5) {};
		\node [style=none] (6) at (-1, -1.75) {\scriptsize{\texttt{input}}};
		\node [style=none] (7) at (-3.75, 3.15) {};
		\node [style=none] (8) at (4, 3.15) {};
		\node [style=none] (9) at (-3.75, -2.25) {};
		\node [style=none] (10) at (4, -2.25) {};
		\node [style=none] (11) at (-3.25, 3.5) {\texttt{net}};
		\node [style=none] (14) at (1, -1.5) {};
		\node [style=none] (15) at (1, -1.75) {\scriptsize{\texttt{input}}};
		\node [style=leaf ParamNode, minimum width=2cm] (16) at (0, 2) {};
		\node [style=none] (19) at (-1, 2) {};
		\node [style=leaf Node] (21) at (1, 0) {};
		\node [style=none] (22) at (1, 2) {};
		\node [style=none] (23) at (-1, 0.825) {\scriptsize{\texttt{node1}}};
		\node [style=none] (24) at (1, 0.825) {\scriptsize{\texttt{node2}}};
		\node [style=none] (25) at (0, 2.825) {\scriptsize{\texttt{paramnode}}};
	\end{pgfonlayer}
	\begin{pgfonlayer}{edgelayer}
		\draw (0) to (3.center);
		\draw [style={dashed_edge}] (7.center) to (9.center);
		\draw [style={dashed_edge}] (7.center) to (8.center);
		\draw [style={dashed_edge}] (8.center) to (10.center);
		\draw [style={dashed_edge}] (10.center) to (9.center);
		\draw [style=Edge, in=165, out=165, looseness=2.25] (0) to (19.center);
		\draw [style=Edge, in=15, out=15, looseness=2.25] (21) to (22.center);
		\draw [style=Edge] (0) to (21);
		\draw [style=Edge] (21) to (14.center);
	\end{pgfonlayer}
\end{tikzpicture}
\end{minipage}\\
In the graphical notation we will use squares for denoting trainable nodes.\\[5pt]
\textbf{Leaf nodes:} These are the nodes that form the \texttt{TensorNetwork}, along with the data nodes (see below).
Usually, leaf nodes will be the trainable nodes.
All leaf nodes contain PyTorch leaf tensors, this is, tensors that are not generated by operations tracked by PyTorch's automatic differentiation engine.
By default, all nodes are leaf.
In the graphical notation, these are the blue nodes.\\[5pt]
\textbf{Data nodes:} These are nodes that are used to store the tensors coming from input data.
It is possible to instantiate data nodes directly by specifying \texttt{data=True} in the initialization of \texttt{Node}.
However, data nodes will be usually created when specifying where they should be put in the network using \texttt{set\_data\_nodes} (see \Cref{sec:training}).
Graphically, data nodes will be depicted as red circles.

\noindent\begin{minipage}[c]{0.60\textwidth}
\begin{lstlisting}
 data1 = tk.Node(name="data1",
                 shape=(5,),
                 axes_names=("feature",),
                 data=True,
                 network=net)

 data2 = tk.Node(name="data2",
                 shape=(5,),
                 axes_names=("feature",),
                 data=True,
                 network=net)

 # We can add input data tensors to the nodes
 data1.tensor = torch.randn(5)
 data2.tensor = torch.randn(5)

 node1["input"] ^ data1["feature"]
 node2["input"] ^ data2["feature"]
\end{lstlisting}
\end{minipage}
\hfill
\begin{minipage}[c]{0.39\textwidth}
    \flushright
    \begin{tikzpicture}
	\begin{pgfonlayer}{nodelayer}
		\node [style=leaf Node] (0) at (-1, 0) {};
		\node [style=data Node] (3) at (-1, -2) {};
		\node [style=none] (7) at (-3.75, 3.15) {};
		\node [style=none] (8) at (4, 3.15) {};
		\node [style=none] (9) at (-3.75, -3.25) {};
		\node [style=none] (10) at (4, -3.25) {};
		\node [style=none] (11) at (-3.25, 3.5) {\texttt{net}};
		\node [style=data Node] (14) at (1, -2) {};
		\node [style=leaf ParamNode, minimum width=2cm] (16) at (0, 2) {};
		\node [style=none] (19) at (-1, 2) {};
		\node [style=leaf Node] (21) at (1, 0) {};
		\node [style=none] (22) at (1, 2) {};
		\node [style=none] (23) at (-1, 0.825) {\scriptsize{\texttt{node1}}};
		\node [style=none] (24) at (1, 0.825) {\scriptsize{\texttt{node2}}};
		\node [style=none] (25) at (0, 2.825) {\scriptsize{\texttt{paramnode}}};
		\node [style=none] (26) at (-1, -2.925) {\scriptsize{\texttt{data1}}};
		\node [style=none] (27) at (1, -2.925) {\scriptsize{\texttt{data2}}};
	\end{pgfonlayer}
	\begin{pgfonlayer}{edgelayer}
		\draw (0) to (3);
		\draw [style={dashed_edge}] (7.center) to (9.center);
		\draw [style={dashed_edge}] (7.center) to (8.center);
		\draw [style={dashed_edge}] (8.center) to (10.center);
		\draw [style={dashed_edge}] (10.center) to (9.center);
		\draw [style=Edge, in=165, out=165, looseness=2.25] (0) to (19.center);
		\draw [style=Edge, in=15, out=15, looseness=2.25] (21) to (22.center);
		\draw [style=Edge] (0) to (21);
		\draw [style=Edge] (21) to (14);
	\end{pgfonlayer}
\end{tikzpicture}
\end{minipage}\\[5pt]
\textbf{Virtual nodes:} These nodes are internal nodes that \tk{} uses as shortcuts when one wishes to have the same tensor in different nodes, or to process auxiliary information without adding new leaf nodes to the network.
This is useful when defining uniform tensor networks \cite{zauner2018uniform}, as all their nodes simply use a single tensor stored in a virtual node instead of one tensor per node.
Virtual nodes are intended to be mainly internal, but they can be created manually using the argument \texttt{virtual=True}.
This is needed when defining custom uniform architectures beyond those natively implemented, such as \texttt{UMPS}, \texttt{UMPO}, \texttt{UPEPS} or \texttt{UTree}. Virtual nodes are represented by empty nodes with dashed borders.

\noindent\begin{minipage}[c]{0.60\textwidth}
\begin{lstlisting}
 virtual = tk.Node(name="virtualnode",
                   shape=(2, 5, 2),
                   axes_names=
                       ("left", "input", "right"),
                   virtual=True,
                   network=net,
                   init_method="randn")

 # Put virtual's tensor into node1 and node2
 node1.set_tensor_from(virtual)
 node2.set_tensor_from(virtual)
\end{lstlisting}
\end{minipage}
\hfill
\begin{minipage}[c]{0.39\textwidth}
    \flushright
    \begin{tikzpicture}
	\begin{pgfonlayer}{nodelayer}
		\node [style=leaf Node] (0) at (-3, 0) {};
		\node [style=data Node] (3) at (-3, -2) {};
		\node [style=none] (7) at (-5.75, 3.15) {};
		\node [style=none] (8) at (4.75, 3.15) {};
		\node [style=none] (9) at (-5.75, -3.25) {};
		\node [style=none] (10) at (4.75, -3.25) {};
		\node [style=none] (11) at (-5.25, 3.5) {\texttt{net}};
		\node [style=data Node] (14) at (-1, -2) {};
		\node [style=leaf ParamNode, minimum width=2cm] (16) at (-2, 2) {};
		\node [style=none] (19) at (-3, 2) {};
		\node [style=leaf Node] (21) at (-1, 0) {};
		\node [style=none] (22) at (-1, 2) {};
		\node [style=none] (23) at (-3, 0.825) {\scriptsize{\texttt{node1}}};
		\node [style=none] (24) at (-1, 0.825) {\scriptsize{\texttt{node2}}};
		\node [style=none] (25) at (-2, 2.825) {\scriptsize{\texttt{paramnode}}};
		\node [style=none] (26) at (-3, -2.925) {\scriptsize{\texttt{data1}}};
		\node [style=none] (27) at (-1, -2.925) {\scriptsize{\texttt{data2}}};
		\node [style=virtual Node] (28) at (3, 0) {};
		\node [style=none] (29) at (1.5, 0) {};
		\node [style=none] (30) at (4.5, 0) {};
		\node [style=none] (31) at (3, -1.5) {};
		\node [style=none] (32) at (3, 0.825) {\scriptsize{\texttt{virtualnode}}};
		\node [style=none] (33) at (2.5, -0.5) {};
		\node [style=none] (34) at (-0.5, -0.5) {};
		\node [style=none] (35) at (2.5, -0.5) {};
		\node [style=none] (36) at (-2.5, -0.5) {};
	\end{pgfonlayer}
	\begin{pgfonlayer}{edgelayer}
		\draw (0) to (3);
		\draw [style={dashed_edge}] (7.center) to (9.center);
		\draw [style={dashed_edge}] (7.center) to (8.center);
		\draw [style={dashed_edge}] (8.center) to (10.center);
		\draw [style={dashed_edge}] (10.center) to (9.center);
		\draw [style=Edge, in=165, out=165, looseness=2.25] (0) to (19.center);
		\draw [style=Edge, in=15, out=15, looseness=2.25] (21) to (22.center);
		\draw [style=Edge] (0) to (21);
		\draw [style=Edge] (21) to (14);
		\draw [style=Edge] (29.center) to (28);
		\draw [style=Edge] (28) to (31.center);
		\draw [style=Edge] (28) to (30.center);
		\draw [style={dashed_edge}, <-, in=-135, out=-45] (34.center) to (33.center);
		\draw [style={dashed_edge}, <-, in=-135, out=-30] (36.center) to (35.center);
	\end{pgfonlayer}
\end{tikzpicture}
\end{minipage}\\[5pt]
\textbf{Resultant nodes:} The result of a contraction of tensors (stored in their respective nodes) is another tensor that must be stored in a new node.
These nodes are resultant nodes, which coexist in the \texttt{TensorNetwork} object together with the original nodes, inheriting their edges.
This allows for subsequent contraction with other neighbouring nodes.
Resultant nodes are automatically created upon contracting two nodes on a same network, or when tracing the model on its first run (see \Cref{sec:operations:trace}). Resultant nodes are displayed as green circles.

\noindent\begin{minipage}[c]{0.5\textwidth}
\begin{lstlisting}
 # Nodes are contracted along their
 # connected edges with the @ operator
 result = node1 @ node2
\end{lstlisting}
\end{minipage}
\hfill
\begin{minipage}[c]{0.44\textwidth}
    \flushright
    \begin{tikzpicture}
	\begin{pgfonlayer}{nodelayer}
		\node [style=leaf Node] (0) at (-3.75, 0) {};
		\node [style=data Node] (3) at (-3.75, -2) {};
		\node [style=none] (7) at (-6.5, 3.15) {};
		\node [style=none] (8) at (6.25, 3.15) {};
		\node [style=none] (9) at (-6.5, -3.25) {};
		\node [style=none] (10) at (6.25, -3.25) {};
		\node [style=none] (11) at (-6, 3.5) {\texttt{net}};
		\node [style=data Node] (14) at (-1.75, -2) {};
		\node [style=leaf ParamNode, minimum width=2cm] (16) at (-2.75, 2) {};
		\node [style=none] (19) at (-3.75, 2) {};
		\node [style=leaf Node] (21) at (-1.75, 0) {};
		\node [style=none] (22) at (-1.75, 2) {};
		\node [style=none] (23) at (-3.75, 0.875) {\scriptsize{\texttt{node1}}};
		\node [style=none] (24) at (-1.75, 0.875) {\scriptsize{\texttt{node2}}};
		\node [style=none] (25) at (-2.75, 2.825) {\scriptsize{\texttt{paramnode}}};
		\node [style=none] (26) at (-3.75, -2.925) {\scriptsize{\texttt{data1}}};
		\node [style=none] (27) at (-1.75, -2.925) {\scriptsize{\texttt{data2}}};
		\node [style=none] (43) at (-0.25, 0) {};
		\node [style=none] (44) at (2, 0) {};
		\node [style=resultant Node] (45) at (3.25, 0) {};
		\node [style=data Node] (46) at (2.25, -2) {};
		\node [style=data Node] (47) at (4.25, -2) {};
		\node [style=leaf ParamNode, minimum width=2cm] (48) at (3.25, 2) {};
		\node [style=none] (49) at (2.25, 2) {};
		\node [style=none] (51) at (4.25, 2) {};
		\node [style=none] (52) at (3.25, 0.825) {\scriptsize{\texttt{result}}};
	\end{pgfonlayer}
	\begin{pgfonlayer}{edgelayer}
            \node [style=virtual ParamNode, minimum width=2cm, minimum height=0.65cm] (34) at (-2.75, 0) {};
		\draw (0) to (3);
		\draw [style={dashed_edge}] (7.center) to (9.center);
		\draw [style={dashed_edge}] (7.center) to (8.center);
		\draw [style={dashed_edge}] (8.center) to (10.center);
		\draw [style={dashed_edge}] (10.center) to (9.center);
		\draw [style=Edge, in=165, out=165, looseness=2.25] (0) to (19.center);
		\draw [style=Edge, in=15, out=15, looseness=2.25] (21) to (22.center);
		\draw [style=Edge] (0) to (21);
		\draw [style=Edge] (21) to (14);
		\draw [style={dashed_edge}, ->] (43.center) to (44.center);
		\draw (45) to (46);
		\draw [style=Edge, in=180, out=165, looseness=2.75] (45) to (49.center);
		\draw [style=Edge] (45) to (47);
		\draw [style=Edge, in=15, out=0, looseness=3.00] (51.center) to (45);
	\end{pgfonlayer}
\end{tikzpicture}
\end{minipage}

\section{Operations between Nodes}\label{sec:operations}
Operations between \tk{} \texttt{Nodes} are instances of a class, \texttt{Operation}, that is optimized to avoid unnecessary repeating of basic computations (see \Cref{sec:memory} for details).
Operations create resultant nodes that inherit information from the parent nodes.
\tk{} implements many operations that are extensions of those in vanilla PyTorch, like \texttt{permute}, \texttt{tprod} (\texttt{outer} in PyTorch), \texttt{mul}, \texttt{div}, \texttt{add} and \texttt{sub}.
The complete list of implemented operations is available in \href{https://joserapa98.github.io/tensorkrowch/_build/html/operations.html}{the corresponding page of the documentation}.
In addition to these, \tk{} includes implementations of operations that are inherent to tensor networks. These are:\\[5pt]
\textbf{Contract:} Enables contraction of all the edges connecting two nodes, although it is also possible to use it to contract only a selected range of edges.
Furthermore, batch contraction is automatically performed if some edge is of batch type (recall, with the string \texttt{"batch"} in its name).
This is the most basic operation one can use, allowing for the contraction of the whole network just by implementing a contraction path.

\noindent\begin{minipage}[c]{0.60\textwidth}
\begin{lstlisting}
 node1 = tk.randn(shape=(2, 4, 3, 6, 2))
 node2 = tk.randn(shape=(3, 2, 5, 4))

 # Edges can also be accessed by index
 node1[2] ^ node2[0]
 node1[4] ^ node2[1]
 # node1 and node2 will belong to the same
 # network upon connection

 result = node1 @ node2 
 # result will have all the non-contracted
 # edges from node1 and node2
\end{lstlisting}
\end{minipage}
\begin{minipage}[c]{0.39\textwidth}
    \flushright
    \begin{tikzpicture}
	\begin{pgfonlayer}{nodelayer}
		\node [style=leaf Node] (0) at (-3.5, 0) {};
		\node [style=none] (7) at (-5.25, 1.9) {};
		\node [style=none] (8) at (5, 1.9) {};
		\node [style=none] (9) at (-5.25, -1.75) {};
		\node [style=none] (10) at (5, -1.75) {};
		\node [style=none] (11) at (-4.75, 2.25) {\texttt{net}};
		\node [style=leaf Node] (21) at (-1.5, 0) {};
		\node [style=none] (43) at (0, 0) {};
		\node [style=none] (44) at (2.25, 0) {};
		\node [style=resultant Node] (45) at (3.5, 0) {};
		\node [style=none] (54) at (-4.75, 1) {};
		\node [style=none] (55) at (-4.75, -1) {};
		\node [style=none] (56) at (-2.25, -1) {};
		\node [style=none] (57) at (-1.5, 1.5) {};
		\node [style=none] (58) at (-0.5, 1.25) {};
		\node [style=none] (59) at (2.25, 1) {};
		\node [style=none] (60) at (2.25, -1) {};
		\node [style=none] (61) at (4.75, -1) {};
		\node [style=none] (62) at (3.5, 1.5) {};
		\node [style=none] (63) at (4.5, 1.25) {};
	\end{pgfonlayer}
	\begin{pgfonlayer}{edgelayer}
		\draw [style={dashed_edge}] (7.center) to (9.center);
		\draw [style={dashed_edge}] (7.center) to (8.center);
		\draw [style={dashed_edge}] (8.center) to (10.center);
		\draw [style={dashed_edge}] (10.center) to (9.center);
		\draw [style={dashed_edge}, ->] (43.center) to (44.center);
		\draw [style=Edge] (54.center) to (0);
		\draw [style=Edge] (55.center) to (0);
		\draw [style=Edge, bend left=60] (0) to (21);
		\draw [style=Edge] (0) to (56.center);
		\draw [style=Edge, bend right=75, looseness=1.75] (0) to (21);
		\draw [style=Edge] (21) to (57.center);
		\draw [style=Edge] (58.center) to (21);
		\draw [style=Edge] (59.center) to (45);
		\draw [style=Edge] (45) to (60.center);
		\draw [style=Edge] (45) to (61.center);
		\draw [style=Edge] (45) to (62.center);
		\draw [style=Edge] (45) to (63.center);
	\end{pgfonlayer}
\end{tikzpicture}
\end{minipage}\\[5pt]
\textbf{Split:} As an inverse of the \texttt{contract} operation, \texttt{split} factorizes the tensor in a node into two, selecting which edges should go to each resultant node.
Factorization algorithms include singular value and QR decompositions, with additional functionalities such as selecting the desired amount of singular values to keep, modifying the rank of the resultant nodes accordingly.
This can be useful to reduce the bond dimension in cases where it might explode during the contraction algorithm.
Furthermore, by iterative application of \texttt{split}, arbitrary tensors can be decomposed into tensor network formats, which enables defining custom tensorization routines \cite{novikov2015tensorizing, lebedev2014speeding, ma2019tensorized}.

\noindent\begin{minipage}[c]{0.45\textwidth}
\begin{lstlisting}
 node = tk.randn(
     shape=(2, 7, 3, 4),
     axes_names=
         ("left_0", "left_1",
          "right_0", "right_1")
 )

 node_left, node_right = tk.split(
     node,
     ["left_0", "right_0"],
     ["left_1", "right_1"],
     mode="svd",
     rank=5
 )
\end{lstlisting}
\end{minipage}
\begin{minipage}[c]{0.54\textwidth}
    \flushright
    \begin{tikzpicture}
	\begin{pgfonlayer}{nodelayer}
		\node [style=leaf Node] (0) at (-3.75, 0) {};
		\node [style=none] (7) at (-6.75, 1.9) {};
		\node [style=none] (8) at (7.25, 1.9) {};
		\node [style=none] (9) at (-6.75, -1.75) {};
		\node [style=none] (10) at (7.25, -1.75) {};
		\node [style=none] (11) at (-6.25, 2.25) {\texttt{net}};
		\node [style=none] (43) at (-1.5, 0) {};
		\node [style=none] (44) at (0.75, 0) {};
		\node [style=resultant Node] (45) at (2.25, 0) {};
		\node [style=none] (54) at (-5.25, 0.5) {};
		\node [style=none] (55) at (-5.25, -0.5) {};
		\node [style=none] (56) at (-2.25, -0.5) {};
		\node [style=none] (64) at (-2.25, 0.5) {};
		\node [style=none] (65) at (-5.75, 0.8) {\scriptsize{\texttt{left\_0}}};
		\node [style=none] (66) at (-5.75, -0.8) {\scriptsize{\texttt{left\_1}}};
		\node [style=none] (67) at (-1.75, 0.75) {\scriptsize{\texttt{right\_0}}};
		\node [style=none] (68) at (-1.75, -0.8) {\scriptsize{\texttt{right\_1}}};
		\node [style=resultant Node] (69) at (4.25, 0) {};
		\node [style=none] (70) at (2.75, 1.25) {};
		\node [style=none] (71) at (1, 0.75) {};
		\node [style=none] (72) at (5.5, -0.25) {};
		\node [style=none] (73) at (3.75, -1) {};
		\node [style=none] (74) at (0.75, 1.05) {\scriptsize{\texttt{left\_0}}};
		\node [style=none] (75) at (3.25, 1.5) {\scriptsize{\texttt{right\_0}}};
		\node [style=none] (76) at (3.5, -1.25) {\scriptsize{\texttt{left\_1}}};
		\node [style=none] (77) at (6, -0.55) {\scriptsize{\texttt{right\_1}}};
	\end{pgfonlayer}
	\begin{pgfonlayer}{edgelayer}
		\draw [style={dashed_edge}] (7.center) to (9.center);
		\draw [style={dashed_edge}] (7.center) to (8.center);
		\draw [style={dashed_edge}] (8.center) to (10.center);
		\draw [style={dashed_edge}] (10.center) to (9.center);
		\draw [style={dashed_edge}, ->] (43.center) to (44.center);
		\draw [style=Edge] (54.center) to (0);
		\draw [style=Edge] (55.center) to (0);
		\draw [style=Edge] (0) to (56.center);
		\draw [style=Edge] (64.center) to (0);
		\draw [style=Edge] (45) to (69);
		\draw [style=Edge] (70.center) to (45);
		\draw [style=Edge] (45) to (71.center);
		\draw [style=Edge] (73.center) to (69);
		\draw [style=Edge] (69) to (72.center);
	\end{pgfonlayer}
\end{tikzpicture}
\end{minipage}\\[5pt]
\textbf{Stack:} In many situations, stacking a set of tensors into a larger tensor with one extra dimension speeds up operations by allowing for parallelizing contractions (see \Cref{sec:memory} for more details).
Creating such stacks is achieved with \texttt{stack}.
The node resulting from this operation is a special type of node, namely a \texttt{StackNode} or \texttt{ParamStackNode}, that is only intended for internal use.

\noindent\begin{minipage}[c]{0.37\textwidth}
\begin{lstlisting}
 net   = tk.TensorNetwork()
 nodes = [
     tk.randn(
         shape=(2, 4, 2),
         axes_names=("left",
                     "input",
                     "right"),
         network=net
     )
     for _ in range(n_nodes)
 ]
 stacknode = tk.stack(nodes)
\end{lstlisting}
\end{minipage}
\begin{minipage}[c]{0.62\textwidth}
    \flushright
    \resizebox{\textwidth}{!}{
            \begin{tikzpicture}
	\begin{pgfonlayer}{nodelayer}
		\node [style=leaf Node] (0) at (-9.5, 0) {};
		\node [style=none] (7) at (-12, 2.65) {};
		\node [style=none] (8) at (13.5, 2.65) {};
		\node [style=none] (9) at (-12, -4.5) {};
		\node [style=none] (10) at (13.5, -4.5) {};
		\node [style=none] (11) at (-11.5, 3) {\texttt{net}};
		\node [style=none] (43) at (1.5, 0) {};
		\node [style=none] (44) at (3.75, 0) {};
		\node [style=none] (54) at (-11, 0) {};
		\node [style=none] (55) at (-9.5, -1.5) {};
		\node [style=none] (64) at (-8, 0) {};
		\node [style=leaf Node] (78) at (-5.75, 0) {};
		\node [style=none] (79) at (-7.25, 0) {};
		\node [style=none] (80) at (-5.75, -1.5) {};
		\node [style=none] (81) at (-4.25, 0) {};
		\node [style=leaf Node] (82) at (-0.75, 0) {};
		\node [style=none] (83) at (-2.25, 0) {};
		\node [style=none] (84) at (-0.75, -1.5) {};
		\node [style=none] (85) at (0.75, 0) {};
		\node [style=none] (86) at (-3.25, 0) {$\cdots$};
		\node [style=resultant Node] (87) at (9.75, -0.75) {};
		\node [style=none] (88) at (8.25, -0.75) {};
		\node [style=none] (89) at (9.75, -2.25) {};
		\node [style=none] (90) at (11.25, -0.75) {};
		\node [style=resultant Node] (91) at (7.75, 1.25) {};
		\node [style=none] (92) at (6.25, 1.25) {};
		\node [style=none] (93) at (7.75, -0.25) {};
		\node [style=none] (94) at (9.25, 1.25) {};
		\node [style=resultant Node] (95) at (7.5, 1.5) {};
		\node [style=none] (96) at (6, 1.5) {};
		\node [style=none] (97) at (7.5, 0) {};
		\node [style=none] (98) at (9, 1.5) {};
		\node [style=none] (99) at (8.75, 0.5) {$\ddots$};
		\node [style=none] (101) at (4.5, 0) {};
		\node [style=none] (102) at (6, 0) {};
		\node [style=none] (103) at (11.5, 0) {};
		\node [style=none] (104) at (13, 0) {};
		\node [style=none] (105) at (8.75, -2.25) {};
		\node [style=none] (106) at (8.75, -3.75) {};
		\node [style=none] (107) at (6, -2.5) {};
		\node [style=none] (108) at (5, -3.5) {};
		\node [style=none] (109) at (-10.75, 0.3) {\scriptsize{\texttt{left}}};
		\node [style=none] (110) at (-8.15, 0.3) {\scriptsize{\texttt{right}}};
		\node [style=none] (111) at (-9.5, -1.95) {\scriptsize{\texttt{input}}};
		\node [style=none] (113) at (5, 0.3) {\scriptsize{\texttt{left}}};
		\node [style=none] (114) at (12.35, 0.3) {\scriptsize{\texttt{right}}};
		\node [style=none] (115) at (8.75, -4.2) {\scriptsize{\texttt{input}}};
		\node [style=none] (116) at (4.25, -3.7) {\scriptsize{\texttt{stack}}};
            \draw [style=Edge] (96.center) to (95);
		\draw [style=Edge] (97.center) to (95);
		\draw [style=Edge] (98.center) to (95);
	\end{pgfonlayer}
	\begin{pgfonlayer}{edgelayer}
            \draw [style=Edge] (101.center) to (102.center);
		\draw [style=Edge] (103.center) to (104.center);
		\draw [style=Edge] (105.center) to (106.center);
		\draw [style=Edge] (108.center) to (107.center);
            \node [style=leaf ParamNode, minimum width=2.75cm, minimum height=2.4cm, fill=white, draw={rgb,255: red,132; green,220; blue,198}, line width=1.5pt] (100) at (8.75, -0.1) {};
		\draw [style={dashed_edge}] (7.center) to (9.center);
		\draw [style={dashed_edge}] (7.center) to (8.center);
		\draw [style={dashed_edge}] (8.center) to (10.center);
		\draw [style={dashed_edge}] (10.center) to (9.center);
		\draw [style={dashed_edge}, ->] (43.center) to (44.center);
		\draw [style=Edge] (54.center) to (0);
		\draw [style=Edge] (55.center) to (0);
		\draw [style=Edge] (64.center) to (0);
		\draw [style=Edge] (79.center) to (78);
		\draw [style=Edge] (80.center) to (78);
		\draw [style=Edge] (81.center) to (78);
		\draw [style=Edge] (83.center) to (82);
		\draw [style=Edge] (84.center) to (82);
		\draw [style=Edge] (85.center) to (82);
		\draw [style=Edge] (88.center) to (87);
		\draw [style=Edge] (89.center) to (87);
		\draw [style=Edge] (90.center) to (87);
		\draw [style=Edge] (92.center) to (91);
		\draw [style=Edge] (93.center) to (91);
		\draw [style=Edge] (94.center) to (91);
	\end{pgfonlayer}
\end{tikzpicture}
    }
\end{minipage}\\[5pt]
\textbf{Unbind:} As the counterpart of \texttt{stack}, this enables one to unbind a stack of resultant nodes, returning them in a list. Only \texttt{StackNode}s and \texttt{ParamStackNode}s can be unbound.

\noindent\begin{minipage}[c]{0.36\textwidth}
\begin{lstlisting}
 stacknode = tk.stack(nodes)
 result = tk.unbind(stacknode)
\end{lstlisting}
\end{minipage}
\begin{minipage}[c]{0.63\textwidth}
    \flushright
    \resizebox{\textwidth}{!}{
        \begin{tikzpicture}
	\begin{pgfonlayer}{nodelayer}
		\node [style=resultant Node] (0) at (2.75, 0) {};
		\node [style=none] (7) at (-12, 2.65) {};
		\node [style=none] (8) at (13.5, 2.65) {};
		\node [style=none] (9) at (-12, -4.5) {};
		\node [style=none] (10) at (13.5, -4.5) {};
		\node [style=none] (11) at (-11.5, 3) {\texttt{net}};
		\node [style=none] (43) at (-1.55, 0) {};
		\node [style=none] (44) at (0.7, 0) {};
		\node [style=none] (54) at (1.25, 0) {};
		\node [style=none] (55) at (2.75, -1.5) {};
		\node [style=none] (64) at (4.25, 0) {};
		\node [style=resultant Node] (78) at (6.5, 0) {};
		\node [style=none] (79) at (5, 0) {};
		\node [style=none] (80) at (6.5, -1.5) {};
		\node [style=none] (81) at (8, 0) {};
		\node [style=resultant Node] (82) at (11.5, 0) {};
		\node [style=none] (83) at (10, 0) {};
		\node [style=none] (84) at (11.5, -1.5) {};
		\node [style=none] (85) at (13, 0) {};
		\node [style=none] (86) at (9, 0) {$\cdots$};
		\node [style=leaf Node] (87) at (-5.5, -0.75) {};
		\node [style=none] (88) at (-7, -0.75) {};
		\node [style=none] (89) at (-5.5, -2.25) {};
		\node [style=none] (90) at (-4, -0.75) {};
		\node [style=leaf Node] (91) at (-7.5, 1.25) {};
		\node [style=none] (92) at (-9, 1.25) {};
		\node [style=none] (93) at (-7.5, -0.25) {};
		\node [style=none] (94) at (-6, 1.25) {};
		\node [style=leaf Node] (95) at (-7.75, 1.5) {};
		\node [style=none] (96) at (-9.25, 1.5) {};
		\node [style=none] (97) at (-7.75, 0) {};
		\node [style=none] (98) at (-6.25, 1.5) {};
		\node [style=none] (99) at (-6.5, 0.5) {$\ddots$};
		\node [style=none] (101) at (-10.75, 0) {};
		\node [style=none] (102) at (-9.25, 0) {};
		\node [style=none] (103) at (-3.75, 0) {};
		\node [style=none] (104) at (-2.25, 0) {};
		\node [style=none] (105) at (-6.5, -2.25) {};
		\node [style=none] (106) at (-6.5, -3.75) {};
		\node [style=none] (107) at (-9.25, -2.5) {};
		\node [style=none] (108) at (-10.25, -3.5) {};
		\node [style=none] (109) at (1.5, 0.3) {\scriptsize{\texttt{left}}};
		\node [style=none] (110) at (4.1, 0.3) {\scriptsize{\texttt{right}}};
		\node [style=none] (111) at (2.75, -1.95) {\scriptsize{\texttt{input}}};
		\node [style=none] (113) at (-10.25, 0.3) {\scriptsize{\texttt{left}}};
		\node [style=none] (114) at (-2.9, 0.3) {\scriptsize{\texttt{right}}};
		\node [style=none] (115) at (-6.5, -4.2) {\scriptsize{\texttt{input}}};
		\node [style=none] (116) at (-11, -3.7) {\scriptsize{\texttt{stack}}};
            \draw [style=Edge] (96.center) to (95);
		\draw [style=Edge] (97.center) to (95);
		\draw [style=Edge] (98.center) to (95);
	\end{pgfonlayer}
	\begin{pgfonlayer}{edgelayer}
            \draw [style=Edge] (101.center) to (102.center);
		\draw [style=Edge] (103.center) to (104.center);
		\draw [style=Edge] (105.center) to (106.center);
		\draw [style=Edge] (108.center) to (107.center);
            \node [style=leaf ParamNode, minimum width=2.75cm, minimum height=2.4cm, fill=white, draw={rgb,255: red,171; green,196; blue,255}, line width=1.5pt] (100) at (-6.5, -0.1) {};
		\draw [style={dashed_edge}] (7.center) to (9.center);
		\draw [style={dashed_edge}] (7.center) to (8.center);
		\draw [style={dashed_edge}] (8.center) to (10.center);
		\draw [style={dashed_edge}] (10.center) to (9.center);
		\draw [style={dashed_edge}, ->] (43.center) to (44.center);
		\draw [style=Edge] (54.center) to (0);
		\draw [style=Edge] (55.center) to (0);
		\draw [style=Edge] (64.center) to (0);
		\draw [style=Edge] (79.center) to (78);
		\draw [style=Edge] (80.center) to (78);
		\draw [style=Edge] (81.center) to (78);
		\draw [style=Edge] (83.center) to (82);
		\draw [style=Edge] (84.center) to (82);
		\draw [style=Edge] (85.center) to (82);
		\draw [style=Edge] (88.center) to (87);
		\draw [style=Edge] (89.center) to (87);
		\draw [style=Edge] (90.center) to (87);
		\draw [style=Edge] (92.center) to (91);
		\draw [style=Edge] (93.center) to (91);
		\draw [style=Edge] (94.center) to (91);
	\end{pgfonlayer}
\end{tikzpicture}
    }
\end{minipage}\\[5pt]
\textbf{Einsum:} Allows for the implementation of complex contractions along sets of edges.
Instead of contracting along a connected edge, one can define a contraction path following the Einstein summation convention to contract along several edges at once. This operation uses opt\_einsum \cite{daniel2018opt} at its core, making specific checks and simplifications beforehand to adhere to the rules and structures defined in \tk{}.
For example, indices used twice can only correspond to nodes already connected by an edge in the specified axes.
There is a variant of this operation, \texttt{stacked\_einsum}, which allows to use lists of nodes as inputs, to perform \texttt{stack} followed by \texttt{einsum} in the same operation.

\noindent\begin{minipage}[c]{0.6\textwidth}
\begin{lstlisting}
 node1 = tk.randn(shape=(10, 15, 100),
                  axes_names=
                      ("left", "right", "batch"))
 node2 = tk.randn(shape=(15, 7, 100),
                  axes_names=
                      ("left", "right", "batch"))
 node3 = tk.randn(shape=(7, 10, 100),
                  axes_names=
                      ("left", "right", "batch"))

 node1["right"] ^ node2["left"]
 node2["right"] ^ node3["left"]
 node3["right"] ^ node1["left"]

 result = tk.einsum("ijb,jkb,kib->b",
                    node1, node2, node3)
\end{lstlisting}
\end{minipage}
\begin{minipage}[c]{0.39\textwidth}
    \flushright
    \resizebox{\textwidth}{!}{
        \begin{tikzpicture}
	\begin{pgfonlayer}{nodelayer}
		\node [style=resultant Node] (0) at (4.25, 0) {};
		\node [style=none] (7) at (-6.75, 1.9) {};
		\node [style=none] (8) at (5.25, 1.9) {};
		\node [style=none] (9) at (-6.75, -1.75) {};
		\node [style=none] (10) at (5.25, -1.75) {};
		\node [style=none] (11) at (-6.25, 2.25) {\texttt{net}};
		\node [style=none] (43) at (0.95, 0) {};
		\node [style=none] (44) at (3.2, 0) {};
		\node [style=none] (55) at (4.25, -1.5) {};
		\node [style=leaf Node] (95) at (-5, 0) {};
		\node [style=none] (96) at (-6.25, 0) {};
		\node [style=none] (97) at (-5, -1.5) {};
		\node [style=leaf Node] (98) at (-3, 0) {};
		\node [style=none] (111) at (4.5, -1.15) {$_b$};
		\node [style=leaf Node] (114) at (-1, 0) {};
		\node [style=none] (115) at (-3, -1.5) {};
		\node [style=none] (116) at (-1, -1.5) {};
		\node [style=none] (117) at (0.25, 0) {};
		\node [style=none] (118) at (-6.25, 1) {};
		\node [style=none] (119) at (0.25, 1) {};
		\node [style=none] (120) at (-3, 1.35) {$_i$};
		\node [style=none] (121) at (-4, -0.4) {$_j$};
		\node [style=none] (122) at (-4.75, -1.15) {$_b$};
		\node [style=none] (123) at (-2.75, -1.15) {$_b$};
		\node [style=none] (124) at (-0.75, -1.15) {$_b$};
		\node [style=none] (125) at (-2, -0.4) {$_k$};
	\end{pgfonlayer}
	\begin{pgfonlayer}{edgelayer}
		\draw [style={dashed_edge}] (7.center) to (9.center);
		\draw [style={dashed_edge}] (7.center) to (8.center);
		\draw [style={dashed_edge}] (8.center) to (10.center);
		\draw [style={dashed_edge}] (10.center) to (9.center);
		\draw [style={dashed_edge}, ->] (43.center) to (44.center);
		\draw [style=Edge] (55.center) to (0);
		\draw [style=Edge] (96.center) to (95);
		\draw [style=Edge] (97.center) to (95);
		\draw [style=Edge] (98) to (95);
		\draw [style=Edge] (98) to (114);
		\draw [style=Edge] (115.center) to (98);
		\draw [style=Edge] (116.center) to (114);
		\draw [style=Edge] (119.center) to (118.center);
		\draw [style=Edge] (118.center) to (96.center);
		\draw [style=Edge] (117.center) to (119.center);
		\draw [style=Edge] (117.center) to (114);
	\end{pgfonlayer}
\end{tikzpicture}
    }
\end{minipage}

For a comprehensive explanation of all these operations and the arguments they admit, the reader is referred to the \href{https://joserapa98.github.io/tensorkrowch}{\tk{} documentation}.

\subsection{Trace and Reset}\label{sec:operations:trace}
Every operation in \tk{} returns a new resultant \texttt{Node} that stores the output tensor and inherits the non-contracted edges of its parents.
Therefore, in order to reduce the creation of redundant nodes and the amount of memory used for this purpose, it is useful to generate void containers for the resultant tensors.
This is achieved with the \texttt{trace} operation, which can be called using an example input with batch dimension 1 in order to create all the necessary resultant nodes in the fastest way possible.
Further details and explicit comparisons can be found in \Cref{sec:memory}.

\noindent\begin{minipage}[c]{0.60\textwidth}
\begin{lstlisting}
 # Shape: batch_size x n_features x feature_dim
 data = torch.randn(100, 20, 5)
 example = torch.zeros(1, 20, 5)

 net.trace(example)
\end{lstlisting}
\end{minipage}

The inverse of the \texttt{trace} function is \texttt{reset}.
This function deletes all the resultant nodes created during training, resetting the network to its initial state.
This is useful when one wants to make changes to the structure of the network, to switch on/off the memory modes, or to save a trained model (otherwise, calling \texttt{torch.save(net.state\_dict())} will save a \textit{traced} model, whose parameters might not be exactly the same as the ones in the original model, due to internal optimizations).

\noindent\begin{minipage}[c]{0.60\textwidth}
\begin{lstlisting}
 # After training
 net.reset()

 # Save model
 torch.save(net.state_dict(), "net.pt")

 # Load model
 new_net = TensorNetwork()
 new_net.load_state_dict(torch.load("net.pt"))
\end{lstlisting}
\end{minipage}

\section{Building and training tensor network models}\label{sec:training}
\Cref{sec:components,sec:operations} showed how one can create nodes belonging to a \texttt{TensorNetwork}, and operate them to contract the network.
However, although this functionality might be useful for experimentation, the main usage of \tk{} will be to define custom models as subclasses of \texttt{TensorNetwork}.
This allows, for instance, to instantiate tensor networks that work as any other PyTorch layer.

The workflow to define custom tensor networks is similar to how one defines custom layers in PyTorch.
There, one needs to subclass \texttt{torch.nn.Module}, to define the parameters and architecture of the layer in the \texttt{\_\_init\_\_} method, and to specify how input data is processed by the layer in the \texttt{forward} method.

Similarly, for defining a custom tensor network in \tk{} one needs to subclass \texttt{TensorNetwork}, overriding the following methods:\\[5pt]
\textbf{\texttt{\_\_init\_\_}:} Defines the graph of the tensor network and initializes the tensors of the nodes.

\noindent\begin{minipage}[c]{0.65\textwidth}
\begin{lstlisting}
 class CustomNetwork(tk.TensorNetwork):

     def __init__(self):
         super().__init__(name="CustomNetwork")
         
         self.node1 = tk.randn(
             shape=(2, 5, 2),
             axes_names=("left", "input", "right"),
             name="node1",
             network=self
         )
         self.node2 = tk.randn(
             shape=(2, 5, 2),
             axes_names=("left", "input", "right"),
             name="node2",
             network=self
         )
         self.paramnode = tk.randn(
             shape=(2, 2),
             axes_names=("left", "right"),
             name="paramnode",
             network=self,
             param_node=True
         )
                                  
         self.node1["right"] ^ self.node2["left"]
         self.paramnode["left"] ^ self.node1["left"]
         self.paramnode["right"] ^ self.node2["right"]
\end{lstlisting}
\end{minipage}
\begin{minipage}[c]{0.34\textwidth}
    \flushright
    \begin{tikzpicture}
	\begin{pgfonlayer}{nodelayer}
		\node [style=leaf Node] (0) at (-1, 0) {};
		\node [style=none] (3) at (-1, -1.5) {};
		\node [style=none] (6) at (-1, -1.75) {\scriptsize{\texttt{input}}};
		\node [style=none] (7) at (-3.75, 3.15) {};
		\node [style=none] (8) at (4, 3.15) {};
		\node [style=none] (9) at (-3.75, -2.25) {};
		\node [style=none] (10) at (4, -2.25) {};
		\node [style=none] (11) at (-1.675, 3.5) {\small{\texttt{CustomNetwork}}};
		\node [style=none] (14) at (1, -1.5) {};
		\node [style=none] (15) at (1, -1.75) {\scriptsize{\texttt{input}}};
		\node [style=leaf ParamNode, minimum width=2cm] (16) at (0, 2) {};
		\node [style=none] (19) at (-1, 2) {};
		\node [style=leaf Node] (21) at (1, 0) {};
		\node [style=none] (22) at (1, 2) {};
		\node [style=none] (23) at (-1, 0.825) {\scriptsize{\texttt{node1}}};
		\node [style=none] (24) at (1, 0.825) {\scriptsize{\texttt{node2}}};
		\node [style=none] (25) at (0, 2.825) {\scriptsize{\texttt{paramnode}}};
	\end{pgfonlayer}
	\begin{pgfonlayer}{edgelayer}
		\draw (0) to (3.center);
		\draw [style={dashed_edge}] (7.center) to (9.center);
		\draw [style={dashed_edge}] (7.center) to (8.center);
		\draw [style={dashed_edge}] (8.center) to (10.center);
		\draw [style={dashed_edge}] (10.center) to (9.center);
		\draw [style=Edge, in=165, out=165, looseness=2.25] (0) to (19.center);
		\draw [style=Edge, in=15, out=15, looseness=2.25] (21) to (22.center);
		\draw [style=Edge] (0) to (21);
		\draw [style=Edge] (21) to (14.center);
	\end{pgfonlayer}
\end{tikzpicture}
\end{minipage}\\[5pt]
\textbf{\texttt{set\_data\_nodes} (optional):} Creates the data nodes where the data tensors will be placed. Usually, it will just select the edges to which the data nodes should be connected, and call the parent method.

\noindent\begin{minipage}[c]{0.60\textwidth}
\begin{lstlisting}
 class CustomNetwork(tk.TensorNetwork):

     # ...
     
     def set_data_nodes(self):
         # Collect edges to which data
         # nodes will be connected
         input_edges = [self.node1["input"],
                        self.node2["input"]]
         
         # Define number of batch indices
         # for the input
         num_batch_edges = 1
         
         # Call parent method
         super().set_data_nodes(input_edges,
                                num_batch_edges)
\end{lstlisting}
\end{minipage}
\begin{minipage}[c]{0.39\textwidth}
    \flushright
    \begin{tikzpicture}
	\begin{pgfonlayer}{nodelayer}
		\node [style=leaf Node] (0) at (-1, 0) {};
		\node [style=data Node] (3) at (-1, -2) {};
		\node [style=none] (6) at (-1, -3.775) {\scriptsize{\texttt{batch}}};
		\node [style=none] (7) at (-3.75, 3.15) {};
		\node [style=none] (8) at (4, 3.15) {};
		\node [style=none] (9) at (-3.75, -4.25) {};
		\node [style=none] (10) at (4, -4.25) {};
		\node [style=none] (11) at (-1.675, 3.5) {\small{\texttt{CustomNetwork}}};
		\node [style=data Node] (14) at (1, -2) {};
		\node [style=none] (15) at (1, -3.775) {\scriptsize{\texttt{batch}}};
		\node [style=leaf ParamNode, minimum width=2cm] (16) at (0, 2) {};
		\node [style=none] (19) at (-1, 2) {};
		\node [style=leaf Node] (21) at (1, 0) {};
		\node [style=none] (22) at (1, 2) {};
		\node [style=none] (23) at (-1, 0.825) {\scriptsize{\texttt{node1}}};
		\node [style=none] (24) at (1, 0.825) {\scriptsize{\texttt{node2}}};
		\node [style=none] (25) at (0, 2.825) {\scriptsize{\texttt{paramnode}}};
		\node [style=none] (26) at (-1, -3.5) {};
		\node [style=none] (27) at (1, -3.5) {};
	\end{pgfonlayer}
	\begin{pgfonlayer}{edgelayer}
		\draw (0) to (3);
		\draw [style={dashed_edge}] (7.center) to (9.center);
		\draw [style={dashed_edge}] (7.center) to (8.center);
		\draw [style={dashed_edge}] (8.center) to (10.center);
		\draw [style={dashed_edge}] (10.center) to (9.center);
		\draw [style=Edge, in=165, out=165, looseness=2.25] (0) to (19.center);
		\draw [style=Edge, in=15, out=15, looseness=2.25] (21) to (22.center);
		\draw [style=Edge] (0) to (21);
		\draw [style=Edge] (21) to (14);
		\draw (26.center) to (3);
		\draw (27.center) to (14);
	\end{pgfonlayer}
\end{tikzpicture}
\end{minipage}\\[5pt]
\textbf{\texttt{add\_data} (optional):} Places input data into the previously specified data nodes.
Commonly, all data nodes will have the same shape, namely $b \times d$, being $b$ the batch size and $d$ the feature dimension.
Assuming there are $n$ of these nodes (typically one per feature), the input to \texttt{add\_data} must be a tensor of dimension $b \times n \times d$.
On default operation, the input tensor will be then unbound at its second axis, delivering each slice of shape $b \times d$ to each of the data nodes.
However, this method can be overridden to customize how data is set into the data nodes.\\[5pt]
\textbf{\texttt{contract}:} Very much like the \texttt{forward} method in PyTorch, this is the main method that describes how the components of the network are combined.
In contrast to vanilla PyTorch, however, in a \tk{} \texttt{TensorNetwork} the \texttt{forward} method shall not be overriden, since its goal is to just call \texttt{set\_data\_nodes}, if needed, \texttt{add\_data}, and \texttt{contract}, and then it will return the tensor corresponding to the last resultant node.
Instead, one should customize the \texttt{contract} method.
\tk{} does not implement algorithms for searching optimal contraction paths~\cite{gray2021hyper}.
Thus, one must specify custom contraction algorithms for each user-defined tensor network, via \texttt{einsum} (recall \Cref{sec:operations}) or by any other means.
As will be detailed in \Cref{sec:memory}, the order in which \tk{} \texttt{Operations} appear in the algorithm is significant, being mandatory that the last \texttt{Operation} is the one returning the final node.

\noindent\begin{minipage}{0.80\textwidth}
\begin{lstlisting}
 class CustomNetwork(tk.TensorNetwork):
     
     # ...
     
     def contract(self):
         stack_nodes = tk.stack([self.node1, self.node2])
         stack_data  = tk.stack(list(self.data_nodes.values()))
         
         # Stacks need to be reconnected before contraction
         stack_nodes["input"] ^ stack_data["feature"]
         stack_result = stack_nodes @ stack_data
         
         stack_result = tk.unbind(stack_result)
         
         result = stack_result[0] @ stack_result[1]
         
         # Last operation must return the output node
         result @= self.paramnode
         return result
\end{lstlisting}
\end{minipage}
\begin{equation*}
    \resizebox{\textwidth}{!}{
        \begin{tikzpicture}
	\begin{pgfonlayer}{nodelayer}
		\node [style=leaf Node] (0) at (-1, 0) {};
		\node [style=data Node] (3) at (-1, -2) {};
		\node [style=none] (6) at (-1, -3.775) {\scriptsize{\texttt{batch}}};
		\node [style=none] (7) at (-3.75, 3.15) {};
		\node [style=none] (8) at (32, 3.15) {};
		\node [style=none] (9) at (-3.75, -4.25) {};
		\node [style=none] (10) at (32, -4.25) {};
		\node [style=none] (11) at (-1.675, 3.5) {\small{\texttt{CustomNetwork}}};
		\node [style=data Node] (14) at (1, -2) {};
		\node [style=none] (15) at (1, -3.775) {\scriptsize{\texttt{batch}}};
		\node [style=leaf ParamNode, minimum width=2cm] (16) at (0, 2) {};
		\node [style=none] (19) at (-1, 2) {};
		\node [style=leaf Node] (21) at (1, 0) {};
		\node [style=none] (22) at (1, 2) {};
		\node [style=none] (23) at (-1, 0.825) {\scriptsize{\texttt{node1}}};
		\node [style=none] (24) at (1, 0.825) {\scriptsize{\texttt{node2}}};
		\node [style=none] (25) at (0, 2.825) {\scriptsize{\texttt{paramnode}}};
		\node [style=none] (26) at (-1, -3.5) {};
		\node [style=none] (27) at (1, -3.5) {};
		\node [style=none] (28) at (2.25, 0) {};
		\node [style=none] (29) at (4.25, 0) {};
		\node [style=resultant Node] (30) at (18.75, 0) {};
		\node [style=none] (31) at (18.75, -1.5) {};
		\node [style=none] (32) at (20.75, -1.5) {};
		\node [style=leaf ParamNode, minimum width=2cm] (33) at (19.75, 2) {};
		\node [style=none] (34) at (18.75, 2) {};
		\node [style=resultant Node] (35) at (20.75, 0) {};
		\node [style=none] (36) at (20.75, 2) {};
		\node [style=resultant Node] (41) at (6.5, 0) {};
		\node [style=resultant Node] (42) at (6.5, -2) {};
		\node [style=none] (43) at (5, 0) {};
		\node [style=none] (44) at (8, 0) {};
		\node [style=none] (45) at (6.5, -3.5) {};
		\node [style=none] (46) at (7.5, -3) {};
		\node [style=none] (47) at (7.5, -1) {};
		\node [style=none] (48) at (6.5, 0.875) {\scriptsize{\texttt{stack\_nodes}}};
		\node [style=none] (49) at (4.5, -2) {\scriptsize{\texttt{stack\_data}}};
		\node [style=none] (50) at (8, -1.25) {\scriptsize{\texttt{stack}}};
		\node [style=none] (51) at (8, -3.25) {\scriptsize{\texttt{stack}}};
		\node [style=none] (52) at (8.75, 0) {};
		\node [style=none] (53) at (10.75, 0) {};
		\node [style=resultant Node] (54) at (13, 0) {};
		\node [style=none] (56) at (11.5, 0) {};
		\node [style=none] (57) at (14.5, 0) {};
		\node [style=none] (58) at (13, -1.5) {};
		\node [style=none] (60) at (14, -1) {};
		\node [style=none] (61) at (13, 0.875) {\scriptsize{\texttt{stack\_result}}};
		\node [style=none] (63) at (14.5, -1.25) {\scriptsize{\texttt{stack}}};
		\node [style=none] (64) at (15.25, 0) {};
		\node [style=none] (65) at (17.25, 0) {};
		\node [style=none] (66) at (13, -1.75) {\scriptsize{\texttt{batch}}};
		\node [style=none] (67) at (6.5, -3.75) {\scriptsize{\texttt{batch}}};
		\node [style=none] (68) at (20.75, -1.75) {\scriptsize{\texttt{batch}}};
		\node [style=none] (69) at (18.75, -1.75) {\scriptsize{\texttt{batch}}};
		\node [style=none] (70) at (22, 0) {};
		\node [style=none] (71) at (24, 0) {};
		\node [style=resultant Node] (72) at (25.75, 0) {};
		\node [style=none] (74) at (25.75, -1.5) {};
		\node [style=leaf ParamNode, minimum width=2cm] (75) at (25.75, 2) {};
		\node [style=none] (76) at (24.75, 2) {};
		\node [style=none] (77) at (26.75, 2) {};
		\node [style=none] (78) at (25.75, 0.825) {\scriptsize{\texttt{result}}};
		\node [style=none] (79) at (25.75, -1.75) {\scriptsize{\texttt{batch}}};
		\node [style=none] (80) at (27.5, 0) {};
		\node [style=none] (81) at (29.5, 0) {};
		\node [style=resultant Node] (82) at (30.5, 0) {};
		\node [style=none] (83) at (30.5, -1.5) {};
		\node [style=none] (84) at (30.5, -1.75) {\scriptsize{\texttt{batch}}};
	\end{pgfonlayer}
	\begin{pgfonlayer}{edgelayer}
		\draw (0) to (3);
		\draw [style={dashed_edge}] (7.center) to (9.center);
		\draw [style={dashed_edge}] (7.center) to (8.center);
		\draw [style={dashed_edge}] (8.center) to (10.center);
		\draw [style={dashed_edge}] (10.center) to (9.center);
		\draw [style=Edge, in=165, out=165, looseness=2.25] (0) to (19.center);
		\draw [style=Edge, in=15, out=15, looseness=2.25] (21) to (22.center);
		\draw [style=Edge] (0) to (21);
		\draw [style=Edge] (21) to (14);
		\draw (26.center) to (3);
		\draw (27.center) to (14);
		\draw [style={dashed_edge}, ->] (28.center) to (29.center);
		\draw (30) to (31.center);
		\draw [style=Edge, in=165, out=165, looseness=2.25] (30) to (34.center);
		\draw [style=Edge, in=15, out=15, looseness=2.25] (35) to (36.center);
		\draw [style=Edge] (30) to (35);
		\draw [style=Edge] (35) to (32.center);
		\draw [style=Edge] (41) to (42);
		\draw [style=Edge] (43.center) to (41);
		\draw [style=Edge] (41) to (44.center);
		\draw [style=Edge] (45.center) to (42);
		\draw [style=Edge] (46.center) to (42);
		\draw [style=Edge] (41) to (47.center);
		\draw [style={dashed_edge}, ->] (52.center) to (53.center);
		\draw [style=Edge] (56.center) to (54);
		\draw [style=Edge] (54) to (57.center);
		\draw [style=Edge] (54) to (60.center);
		\draw [style=Edge] (58.center) to (54);
		\draw [style={dashed_edge}, ->] (64.center) to (65.center);
		\draw [style={dashed_edge}, ->] (70.center) to (71.center);
		\draw [style=Edge, in=180, out=165, looseness=2.75] (72) to (76.center);
		\draw [style=Edge] (72) to (74.center);
		\draw [style=Edge, in=15, out=0, looseness=3.00] (77.center) to (72);
		\draw [style={dashed_edge}, ->] (80.center) to (81.center);
		\draw [style=Edge] (82) to (83.center);
	\end{pgfonlayer}
\end{tikzpicture}
    }
\end{equation*}

With the subclass correctly defined, one can now instantiate the custom network and feed  it with new input data tensors:

\noindent\begin{minipage}[c]{0.55\textwidth}
\begin{lstlisting}
 net = CustomNetwork() 

 # Pass data to the model
 # Shape:
 #    batch_size x n_features x feature_dim
 data = torch.randn(100, 2, 5)
 result = net(data)
\end{lstlisting}
\end{minipage}
\begin{minipage}[c]{0.44\textwidth}
    \flushright
    \resizebox{\textwidth}{!}{
        \begin{tikzpicture}
	\begin{pgfonlayer}{nodelayer}
		\node [style=leaf Node] (0) at (-1, 0) {};
		\node [style=data Node] (3) at (-1, -2) {};
		\node [style=none] (6) at (-1, -3.775) {\scriptsize{\texttt{batch}}};
		\node [style=none] (7) at (-6.5, 3.15) {};
		\node [style=none] (8) at (9.75, 3.15) {};
		\node [style=none] (9) at (-6.5, -4.25) {};
		\node [style=none] (10) at (9.75, -4.25) {};
		\node [style=none] (11) at (-4.425, 3.5) {\small{\texttt{CustomNetwork}}};
		\node [style=data Node] (14) at (1, -2) {};
		\node [style=none] (15) at (1, -3.775) {\scriptsize{\texttt{batch}}};
		\node [style=leaf ParamNode, minimum width=2cm] (16) at (0, 2) {};
		\node [style=none] (19) at (-1, 2) {};
		\node [style=leaf Node] (21) at (1, 0) {};
		\node [style=none] (22) at (1, 2) {};
		\node [style=none] (23) at (-1, 0.825) {\scriptsize{\texttt{node1}}};
		\node [style=none] (24) at (1, 0.825) {\scriptsize{\texttt{node2}}};
		\node [style=none] (25) at (0, 2.825) {\scriptsize{\texttt{paramnode}}};
		\node [style=none] (26) at (-1, -3.5) {};
		\node [style=none] (27) at (1, -3.5) {};
		\node [style=none] (28) at (2.25, 0) {};
		\node [style=none] (29) at (4.25, 0) {};
		\node [style=none] (80) at (5.75, 0) {};
		\node [style=none] (81) at (7.75, 0) {};
		\node [style=resultant Node] (82) at (8.75, 0) {};
		\node [style=none] (83) at (8.75, -1.5) {};
		\node [style=none] (84) at (8.75, -1.75) {\scriptsize{\texttt{batch}}};
		\node [style=none] (85) at (5, 0) {$\cdots$};
		\node [style=none] (86) at (-5.25, -0.5) {};
		\node [style=leaf Node] (87) at (-5.25, -2) {};
		\node [style=none] (88) at (-5.25, -3.775) {\scriptsize{\texttt{batch}}};
		\node [style=leaf Node] (89) at (-3.75, -2) {};
		\node [style=none] (90) at (-3.75, -3.775) {\scriptsize{\texttt{batch}}};
		\node [style=none] (91) at (-3.75, -0.5) {};
		\node [style=none] (92) at (-5.25, -3.5) {};
		\node [style=none] (93) at (-3.75, -3.5) {};
		\node [style=none] (94) at (-4.5, 0) {\small{New data}};
		\node [style=none] (95) at (-3.125, -2.525) {};
		\node [style=none] (96) at (-1.65, -2.5) {};
		\node [style=none] (97) at (-4.7, -2.55) {};
		\node [style=none] (98) at (0.45, -2.55) {};
	\end{pgfonlayer}
	\begin{pgfonlayer}{edgelayer}
		\draw (0) to (3);
		\draw [style={dashed_edge}] (7.center) to (9.center);
		\draw [style={dashed_edge}] (7.center) to (8.center);
		\draw [style={dashed_edge}] (8.center) to (10.center);
		\draw [style={dashed_edge}] (10.center) to (9.center);
		\draw [style=Edge, in=165, out=165, looseness=2.25] (0) to (19.center);
		\draw [style=Edge, in=15, out=15, looseness=2.25] (21) to (22.center);
		\draw [style=Edge] (0) to (21);
		\draw [style=Edge] (21) to (14);
		\draw (26.center) to (3);
		\draw (27.center) to (14);
		\draw [style={dashed_edge}, ->] (28.center) to (29.center);
		\draw [style={dashed_edge}, ->] (80.center) to (81.center);
		\draw [style=Edge] (82) to (83.center);
		\draw (86.center) to (87);
		\draw [style=Edge] (91.center) to (89);
		\draw (92.center) to (87);
		\draw (93.center) to (89);
		\draw [style={dashed_edge}, ->, in=-150, out=-45, looseness=1.25] (95.center) to (96.center);
		\draw [style={dashed_edge}, ->, in=-150, out=-45] (97.center) to (98.center);
	\end{pgfonlayer}
\end{tikzpicture}
    }
\end{minipage}\\[3pt]

As mentioned in the beginning of the section, creating a custom network as a subclass of \texttt{TensorNetwork} makes the integration of tensor network layers within PyTorch models straightforward:

\noindent\begin{minipage}[c]{0.60\textwidth}
\begin{lstlisting}
 import torch.nn as nn

 # Combine built-in custom TN layer with PyTorch layers
 model = nn.Sequential(
     tk.models.MPSLayer(n_features=100,
                        in_dim=2,
                        out_dim=10,
                        bond_dim=5),
     nn.ReLU(),
     nn.Linear(10, 10)
 )
\end{lstlisting}
\end{minipage}

The last codeblock contains a built-in class, \texttt{MPSLayer}, that readily implements a MPS architecture.
Similar implementations are available for MPOs, PEPS and TTNs, both uniform and non-uniform. This direct integration also enables to efficiently define tensorized neural network models through the interleaving of MPOs, or simply \texttt{MPSLayers}, with the common nonlinearities of neural networks.

In general, tensor networks have gauge symmetries, i.e., several collections of parameters describe the exact same final tensor.
In certain tensor network architectures there exist ways to define a preferred set of parameters.
This is known as choosing a \textit{canonical form} \cite{perez2006matrix,acuaviva2022minimal}, and is desirable in some physics applications.
In machine learning, canonical forms have been associated to benefits in terms of privacy preservation \cite{pozas2022physics}.
\tk{} allows, in the built-in MPS and TTN implementations, to compute canonical forms using the function \texttt{canonicalize} (or \texttt{canonicalize\_univocal} for the univocal canonical form described in~\cite{pozas2022physics}).

\subsection{Optimization}
The provided code enables users to create custom \texttt{TensorNetwork} models. To train them, different approaches can be followed. In PyTorch, it is customary to pass the model parameters to native optimizers that implement different gradient descent algorithms. Gradients are automatically computed via the automatic differentiation engine, which tracks all operations in which model parameters are involved and computes gradients by applying the chain rule. Since \texttt{TensorNetwork} models are defined as subclasses of \texttt{torch.nn.Module}, and the \texttt{ParamNodes} within these are of type \texttt{torch.nn.Parameter}, this type of gradient-based optimizations are effortlessly implemented to train the \texttt{ParamNodes}. This approach enables the implementation of models similar to the one presented in \cite{novikov2016exponential}, where all MPS nodes are optimized at the same time.

This, however, is not the only way to train tensor networks in \tk{}. With the capability to parameterize and de-parameterize \texttt{Nodes} and \texttt{ParamNodes}, respectively, and assign different tensors to specific \texttt{Nodes}, various optimization schemes can be explored. For instance, DMRG-like approaches \cite{stoudenmire2016supervised} can be easily implemented by freezing all MPS nodes, except for a trainable block that can traverse the matrix chain.

The \tk{} documentation includes \href{https://joserapa98.github.io/tensorkrowch/_build/html/examples.html}{examples} illustrating how to train MPS models in different fashions, including the one in \cite{novikov2016exponential}, as well as DMRG approaches as explained in \cite{stoudenmire2016supervised}. Besides those, implementations of hybrid tensorial neural network models \cite{glasser2020probabilistic} and tensorized neural networks \cite{novikov2015tensorizing} can be found.

\section{Time and memory optimizations}\label{sec:memory}
Operating tensor networks requires a careful handling of memory, since the memory requirement may vary drastically with the contraction path.
In addition to this, it is always desirable to have fast and efficient operations in machine learning pipelines.
To ensure efficient memory utilization, \tk{} employs a memory management scheme where nodes do not possess their own memory. Instead, the memory is stored within the \texttt{TensorNetwork} object, and nodes are just pointers to the corresponding addresses in this shared memory.
This design choice enables memory sharing among all elements of the model, facilitating operations and allowing nodes to utilize tensors from other nodes.
However, it is important to note that memory sharing is limited to elements within the same object, meaning that nodes created in different \texttt{TensorNetwork}s will not share memory.
By adopting this memory management approach, \tk{} incorporates a range of optimizations that effectively reduce both time and memory overheads:

\subsection{Operations}
Tensor network operations in \tk{} are not simple functions, but rather instances of a class, \texttt{Operation}, that is designed to minimize redundant steps during training. Each node operation consists of two functions: one that is executed the first time the operation is called, and one that is executed in every subsequent call with the same arguments.
Although these functions are similar, the former makes extra computations regarding the creation of the resultant nodes and some auxiliary operations that yield the same result in every call.
For instance, when contracting two nodes, tensors are typically permuted first; how this permutation is carried out is always the same, in spite of the fact that the tensors themselves may be different.

Furthermore, to keep track of repeated calls to an \texttt{Operation}, a new object is created during the first run: \texttt{Successor}.
This is a class intended for internal use that acts as a cache memory, storing the arguments that were used to call the operation, some hints regarding the auxiliary tensor-network-related computations, and a reference to the resultant nodes created.
Hence, once an operation has been called, both the parent and children nodes are determined, and only their tensors will change in further contractions. 
This enables to reduce all the code of the contraction algorithm, which may include plain Python code to collect parent nodes, into a sequence of calls to \tk{} \texttt{Operations}.
Because of this simplification, the order in which these operations are called is relevant.
Consequently, the last operation must always be the one returning the final node that corresponds to the contraction of the entire network.

These two optimizations (having different functions for different calls, and exploiting cache memory) break the whole contraction into a set of basic tensor operations that are computed sequentially, thus improving the efficiency of the training process.

\subsection{Trace}
\begin{figure}[h]
    \includegraphics[width=0.42\textwidth]{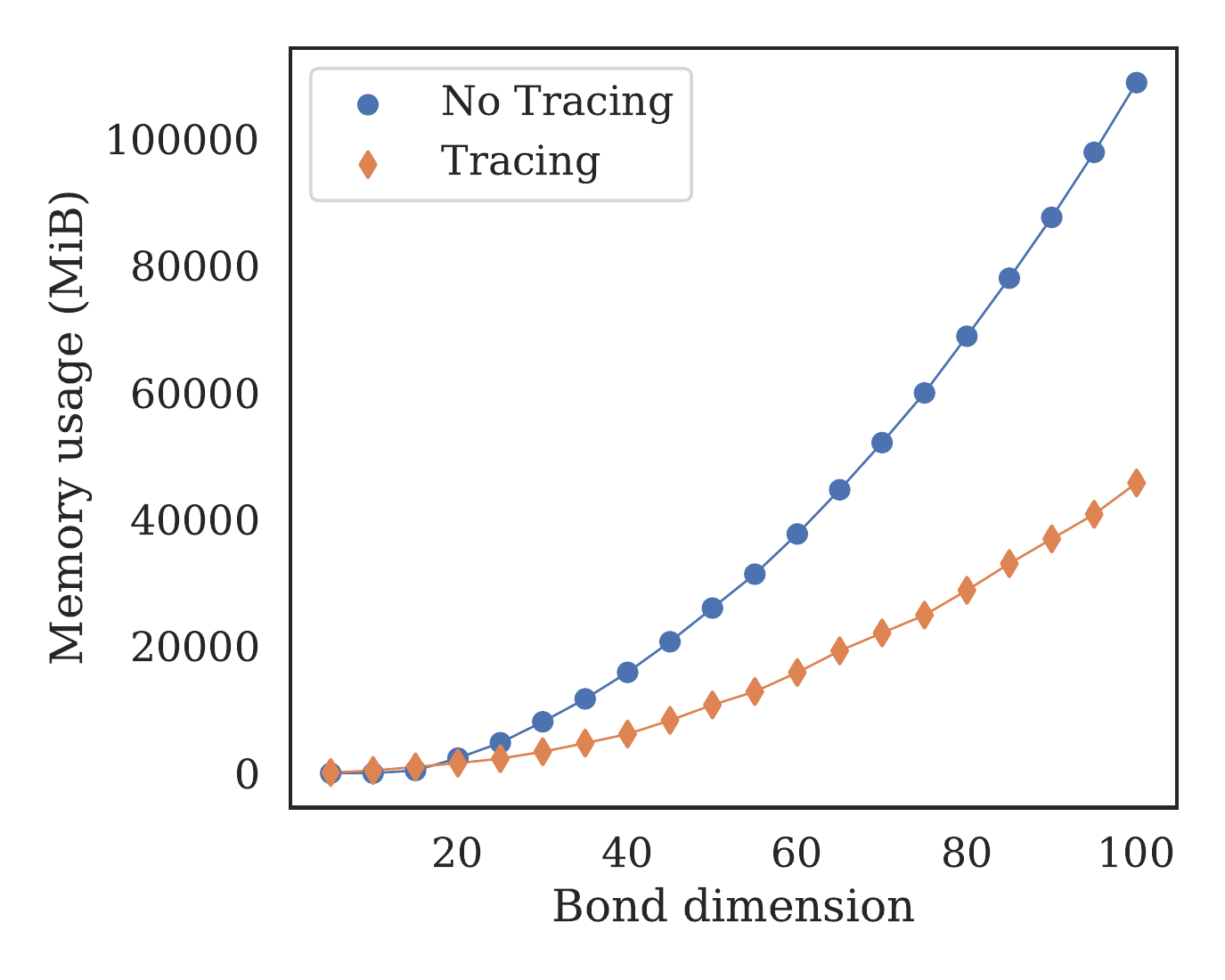}
    \centering
    \caption{
        Comparison of the maximum memory usage of one contraction of the built-in \texttt{MPSLayer} tensor network, when the model is traced or not, using different bond dimensions.
        Contraction is performed in a training regime: 1) an example data tensor is passed through the model, 2) gradients are computed via backpropagation, and 3) parameters are updated according to the gradients.
        All the contractions are computed in CPU using a batch size of 500, with both memory modes \texttt{auto\_stack} and \texttt{auto\_unbind} set to \texttt{True}, both contraction arguments \texttt{inline\_input} and \texttt{inline\_mats} also set to \texttt{True}, and the following arguments for the \texttt{MPSLayer} model: \texttt{n\_features}=1000, \texttt{in\_dim}=2, \texttt{out\_dim}=10.
        All the experiments were run on an Intel Xeon CPU E5-2620 v4 with 256GB of RAM.}
    \label{fig:trace}
\end{figure}

In \Cref{sec:operations:trace} the \texttt{trace} operation was introduced as a means of keeping heavy auxiliary operations involved in the first run of the contraction out of the training loop. Tracing the model not only saves time, but also saves memory.
While tracing a \texttt{TensorNetwork} model, a new memory is created to keep track of which nodes are involved in each operation of the contraction algorithm.
This enables to free up the memory of data or resultant nodes that have already taken part in some operation but are not going to be needed any more in the contraction.
An explicit example of the difference in memory usage that tracing the model produces can be found in \Cref{fig:trace}.

\subsection{Memory Modes}
\begin{figure}[t]
    \includegraphics[width=\textwidth]{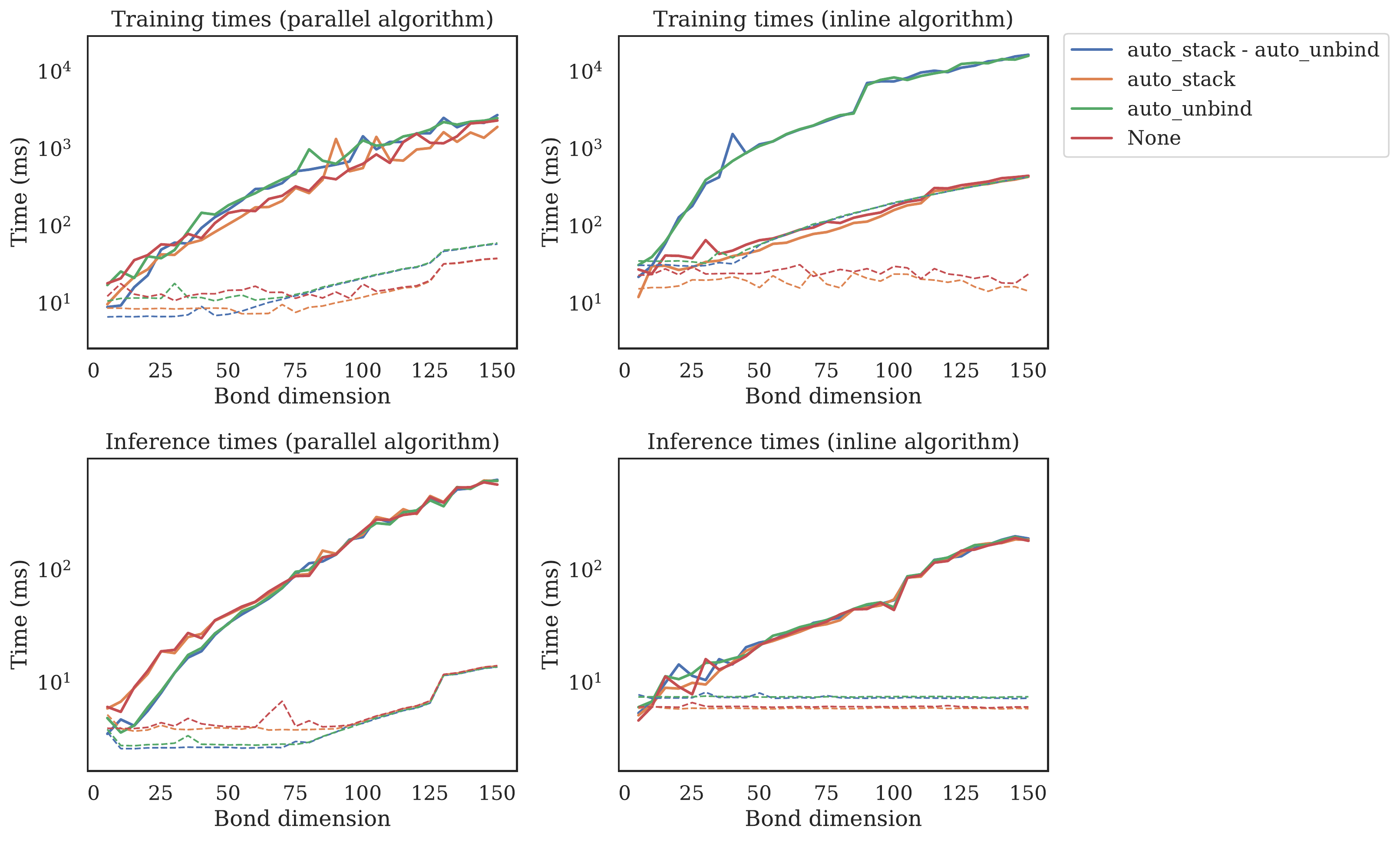}
    \centering
    \caption{
        Comparison of running times of one contraction of the built-in \texttt{MPSLayer} tensor network, using different bond dimensions.
        Contraction is performed in different regimes: training/inference, parallel/inline algorithm, CPU/GPU execution, and using different combinations for the options \texttt{auto\_stack} and \texttt{auto\_unbind}.
        For training, 1) an example data tensor is passed through the model, 2) gradients are computed via backpropagation and 3) parameters are updated according to the gradients.
        For inference, only the example data tensor is passed to perform one contraction of the model.
        Parallel and inline refer to the two possible algorithms that can be used to contract \texttt{MPSLayer}, specified by the argument \texttt{inline\_mats}. When \texttt{inline\_mats = True} (inline), the matrices of the MPS are contracted sequentially; when \texttt{inline\_mats = False} (parallel), matrices are iteratively stacked, contracted in parallel and unbound until the chain of matrices is reduced to a single one.
        The argument \texttt{inline\_input} is always set to \texttt{False}, indicating that the contraction of the MPS nodes with the respective input nodes is performed in parallel.
        Solid lines represent CPU times, dashed lines represent GPU times.
        All the contractions are computed using a batch size of 100 and the following arguments for the \texttt{MPSLayer} model: \texttt{n\_features}=100, \texttt{in\_dim}=2, \texttt{out\_dim}=10.
        All the experiments were run on an Intel Xeon CPU E5-2620 v4 with 256GB of RAM and an NVIDIA GeForce RTX 3090.}
    \label{fig:memory_modes}
\end{figure}

Additionally, there are two modes that can change how tensor networks utilize their memory. These modes, \texttt{auto\_stack} and \texttt{auto\_unbind}, allow the \texttt{stack} and \texttt{unbind} operations, respectively, to reuse information stored in other parts of the memory instead of recalculating it in every contraction.
This helps accelerating both, training and inference.
To illustrate how these modes affect the efficiency of the tensor network model, \Cref{fig:memory_modes} presents a comparison of running times for the built-in \texttt{MPSLayer} class, when these modes are activated or deactivated. 

\section{Additional library information}\label{sec:additional}

\subsection{Limitations and future extensions}
Tensor networks have shown promise in various fields and possess valuable properties for machine learning.
However, their application in this domain is relatively recent, leaving much to be explored in terms of best practices such as initialization techniques, optimizers, and architectures.
In fact, although \tk{} allows the creation of any tensor network, some limitations exist for using certain graphs as models.
For example, contracting PEPS is known to be \#P-hard~\cite{verstraete2006criticality}, and finding optimal contraction paths in arbitrary graphs is also \#P-hard~\cite{damm2002}.
Therefore, \tk{} is presented as a tool that enables rapid prototyping to explore the best techniques to be used in this domain.

We have discussed the various optimizations carried out by \tk{} to avoid redundant calculations as much as possible, ensuring both efficiency and simplicity.
However, for scenarios where efficiency is paramount, such as quantum computer simulation~\cite{pan2022solving,oh2023gbs} or tensorization of neural networks for edge computing~\cite{hawkins2022towards}, there may be libraries with better resource management.
Nevertheless, \tk{} remains useful even in those cases, as it can be part of a preliminary prototyping step.

Since \tk{} has a strong focus on machine learning, the priority has been to implement tensor networks understood as architectures that parameterize intricate families of functions.
Alternative perspectives, such as considering tensor networks as quantum states, have assumed a secondary role.
Despite this, given the importance that the quantum information perspective can have in machine learning explainability \cite{liu2019machine,tangpanitanon2022explainable,aizpurua2024explainable}, \tk{} supports tensors with complex coefficients and implements the calculation of reduced density matrices and entanglement entropy. Building more elaborated explainability tools, and exploiting \tk{} for quantum simulation (for instance, by leveraging Pytorch's automatic differentiation engine for calculating ground-state energies), will be the first steps to be addressed in the future.

On the other hand, other interesting operations include multiple tensor decompositions known in the numerical linear algebra community~\cite{kolda2009tensor}.
\tk{} allows for automatic splitting of tensors appearing in machine learning (such as those appearing in linear and convolutional layers) into MPO forms via singular value decomposition. Other tensorizations that are relevant in the machine learning literature, such as the canonical polyadic \cite{kargas2021cpd} and the Tucker decompositions \cite{liu2022tucker}, are not yet natively implemented.
This is another upcoming objective to develop.

Finally, there are numerous other methods that will progressively be implemented, such as allowing for the modification of the number of edges a node has through a reshape operation, or incorporating visual feedback for observing the constructed graph, akin to the graphical notation used in this paper.

\subsection{Documentation for \tk{}}
The documentation for \tk{} is available online at \url{https://joserapa98.github.io/tensorkrowch}.
It consists of a comprehensive user's guide and an API glossary.
The user's guide provides detailed information that expands upon the topics covered in this paper.
It includes more information on the installation and in-depth tutorials with examples ranging from the basic usage of \texttt{Nodes} and \texttt{Edges} to building advanced hybrid neural-tensor networks like the one discussed in~\cite{glasser2020probabilistic}. 
The API glossary is automatically generated from the docstrings (formatted comments to
code objects), containing detailed information about the public functions and classes defined in \tk{}.

\subsection{Contribution guidelines}
We welcome contributions to \tk{} from the wider communities interested in integrating tensor networks within machine learning frameworks, and in quantum information theory. Contributions can include feedback about the library, feature requests, bug fixes, or code contributions via pull requests.

Feedback and feature requests can be done by opening an issue on the \tk{} GitHub repository \cite{pareja2023tensorkrowch}.
Bug fixes and other pull requests can be done by forking the \tk{} source code, making changes, and then opening a pull request to the \tk{} GitHub repository.
Pull requests are peer-reviewed by \tk{}’s core developers to provide feedback and/or request changes.

Contributors are expected to adhere to \tk{} development practices including style guidelines and unit tests.
Tests are written with the PyTest Python framework and are implemented outside the module.
To test installation or changes, one can download the source code from the repository, and use standard PyTest functions.
For example, executing the following in a Unix terminal in the test folder runs all the tests:

\noindent\begin{minipage}[c]{0.26\textwidth}
\begin{lstlisting}
 python -m pytest -v
\end{lstlisting}
\end{minipage}

\section{Concluding remarks}\label{sec:conclusions}
Machine learning research relies heavily on rapid prototyping and iteration.
By building on top of PyTorch, \tk{} enables these features for machine learning architectures based on tensor networks.
With it, it is possible to use, off the shelf, standard tensor networks (MPS, MPOs, PEPS and TTNs) either as standalone architectures or as layers in deep networks, as well as defining custom networks.
In the latter case, the user has access to customizing fine details of how the network processes input data, such as specifying which parts of the input are sent to which nodes, or fully customizing the contraction path.
Our aim is that \tk{} contributes to the wide adoption of tensor networks by the machine learning community, and that allows to go beyond (and helps advancing) the body of knowledge generated by the communities working in quantum information theory and quantum many-body physics.

In this work we have described the main logic behind \tk{}, as well as the broad families of cases in which it can be used.
We have also detailed the features of its building blocks, and many optimizations that are performed behind the scenes in order to obtain a fast and efficient operation.
Further information on these topics, as well as end-to-end examples of use with state-of-the-art tensor-network and hybrid architectures in standard datasets can be found in the \href{https://joserapa98.github.io/tensorkrowch}{documentation website}.
We strongly encourage any willing user to contribute to the development of the library via its repository \cite{pareja2023tensorkrowch}.

\begin{acknowledgments}
This work is supported by the Spanish Ministry of Science and Innovation MCIN/AEI/10.13039/501100011033 (CEX2019-000904-S, CEX2019-000904-S-20-4, PRE2020-091917 and PID2020-113523GB-I00), Comunidad de Madrid (QUITEMAD-CM P2018/TCS-4342), Universidad Complutense de Madrid (FEI-EU-22-06), and the CSIC Quantum Technologies Platform PTI-001. This work has been financially supported by the Ministry for Digital Transformation and of Civil Service of the Spanish Government through the QUANTUM ENIA project call - Quantum Spain project, and by the European Union through the Recovery, Transformation and Resilience Plan – NextGenerationEU within the framework of the Digital Spain 2026 Agenda.
\end{acknowledgments}

\hypersetup{urlcolor=black}
\bibliographystyle{apsrev4-2}
\bibliography{biblio}

\end{document}